%% file: benchmark0403.tex
\documentclass{article} 
\usepackage{iclr2025_conference,times}
\iclrfinalcopy
\input{math_commands.tex}

\usepackage{hyperref}
\usepackage{url}
\usepackage{tabularx,array,makecell}
\usepackage{multirow}
\usepackage{pifont}
\newcommand{\cmark}{\textcolor{green!60!black}{\ding{51}}}
\newcommand{\xmark}{\textcolor{red!80!black}{\ding{55}}}
\usepackage{graphicx}
\usepackage{booktabs}
\usepackage{wrapfig}
\usepackage{caption}
\usepackage{float}
\usepackage{graphicx}
\usepackage{booktabs}
\usepackage{longtable}
\usepackage{multirow}
\usepackage{tabularx}
\usepackage{booktabs}
\usepackage{array}
\usepackage{threeparttable}
\definecolor{medalgold}{RGB}{212,175,55}
\definecolor{medalsilver}{RGB}{160,160,160}
\definecolor{medalbronze}{RGB}{205,127,50}
\newcommand{\medal}[2]{\textcolor{#1}{\underline{\textbf{#2}}}}

\usepackage{verbatim}
\usepackage{tabularx}
\usepackage{array}
\usepackage{ragged2e} 

\usepackage{setspace}
\usepackage{enumitem}
\usepackage{mdframed}

\usepackage{placeins}

\usepackage{tcolorbox}
\usepackage{fvextra}   
\tcbuselibrary{breakable}

\newtcolorbox{promptbox}{
  breakable,
  enhanced,
  colback=white, colframe=black,
  boxrule=0.6pt, arc=0pt,
  left=2mm, right=2mm, top=1mm, bottom=1mm,
  before skip=0.7\baselineskip,   
  after skip=0.8\baselineskip,    
}
\hypersetup{
    colorlinks=true,
    urlcolor=cyan!60!blue
}

\newcolumntype{Y}{>{\centering\arraybackslash}X}
\newcolumntype{L}[1]{>{\raggedright\arraybackslash}m{#1}}

\title{IMUG-Bench: Benchmarking Unified Multimodal Models on Interleaved Understanding and Generation}

\author{
\makebox[\textwidth][c]{\textbf{
Lingyi Meng$^{*1}$ \quad
Zecong Tang$^{*\dagger 1}$ \quad
Haoran Li$^{*1}$ \quad
Tengju Ru$^{*1}$ \quad
Zhejun Cui$^{1}$
}}\\
\makebox[\textwidth][c]{\textbf{
Weitong Lian$^{1}$ \quad
Qi Kang$^{1}$ \quad
Hangshuo Cao$^{1}$ \quad
Yichen Zhu$^{1}$ \quad
Yechi Liu$^{3}$
}}\\
\makebox[\textwidth][c]{\textbf{
Kaixuan Wang$^{2}$\quad
Yu-Jie Yuan$^{4}$ \quad
Chunwei Wang$^{4}$ \quad
Yu Zhang$^{\ddagger 1}$ \quad
Bo Dai$^{\ddagger 2}$
}}\\[0.5em]
\makebox[\textwidth][c]{$^{1}$ Zhejiang University \quad $^{2}$ The University of Hong
Kong  }\\
\makebox[\textwidth][c]{$^{3}$Institute of Automation, Chinese Academy of Sciences\quad $^{4}$Huawei}\\
\makebox[\textwidth][c]{$^{*}$ Equal contribution\quad $^{\dagger}$ Project leader\quad $^{\ddagger}$ Corresponding author}\\
\\
\makebox[\textwidth][c]{\href{https://github.com/ccccEsion/IMUG-Bench}{\textcolor{cyan!60!blue}{\texttt{https://github.com/ccccEsion/IMUG-Bench}}}}
}

\begin{document}

\maketitle

\begin{abstract}
In recent years, unified multimodal models (UMMs) have emerged to support both understanding and generation within a single framework.
Mastering dynamic, multi-turn interleaved image--text dialogues is a crucial task for UMMs in real-world applications.
However, existing benchmarks fail to evaluate this important task, as they are often limited to single-turn or static settings, and typically overlook exposure bias in multi-turn interactions.
To bridge this gap, we propose IMUG-Bench, a comprehensive benchmark for multi-turn interleaved image--text dialogue of UMMs that jointly evaluates their understanding and generation capabilities.
Our IMUG-Bench comprises three classes: Static Spatial, Temporal Causal, and Hybrid—covering 3,113 samples and 12,034 interaction turns.
It also includes dynamic understanding questions, thereby supporting evaluation that better reflects real-world multi-turn interaction scenarios.
Large-scale experiments on IMUG-Bench systematically evaluate mainstream open-source and closed-source UMMs, revealing their capability boundaries and failure modes, and uncovering pronounced exposure bias on the generation side in multi-turn interactions. 
We further explore several test-time scaling strategies, including Chain-of-Thought, Self-Verification, and Best-of-$N$ Sampling, which effectively improve generation accuracy and mitigate exposure bias in generation tasks. 
These findings provide insights into enhancing the robustness and multi-turn interaction capability of future UMMs.
\end{abstract}

\FloatBarrier
\begin{figure*}[h]
    \centering
    \includegraphics[width=1\linewidth]{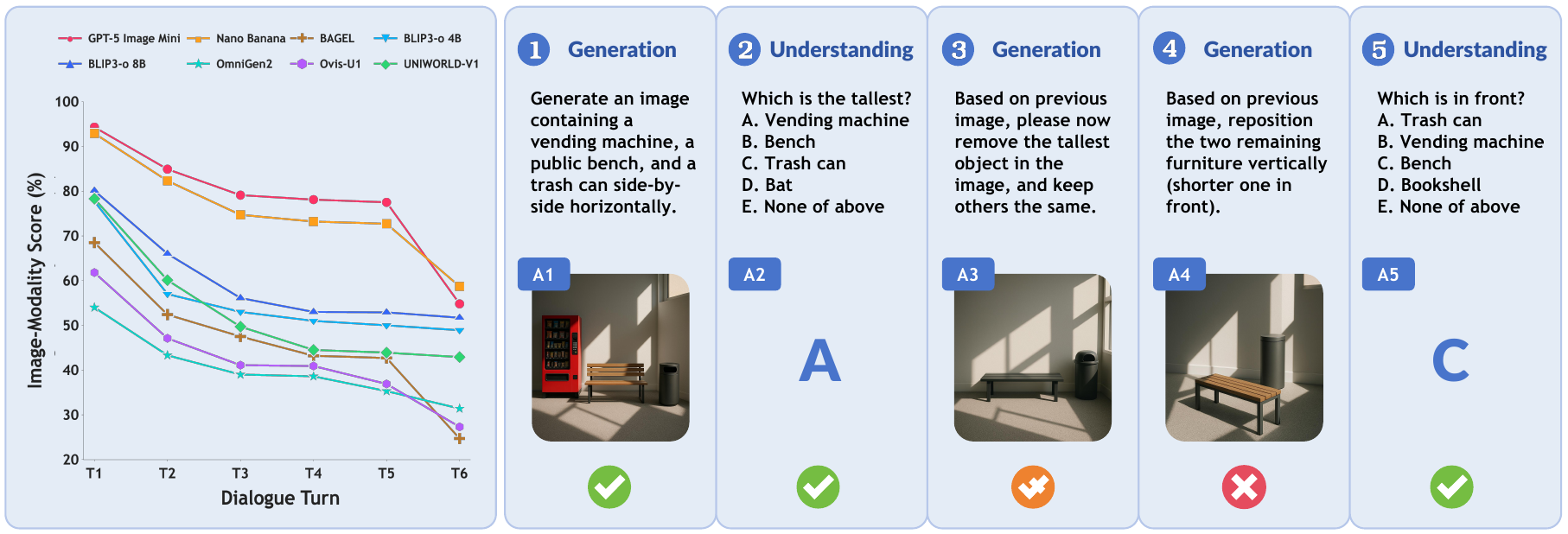}
    \caption{Exposure bias in multi-turn interleaved generation. Left: Image modality scores decline across dialogue turns. Right: An example where image generation quality progressively deteriorates while image understanding remains correct.}
    \label{fig:placeholder}
\end{figure*}
\FloatBarrier

\section{Introduction}
In recent years, multimodal learning models have significantly improved in both understanding and generation tasks
\citep{achiam2023gpt,wu2023multimodal,wang2024qwen2,chen2024internvl,yin_survey_2024}.
On the one hand, understanding models have undergone a paradigm shift from large language models (LLMs) to multimodal large language models (MLLMs), gaining strong visual perception, logical reasoning, and zero-shot generalization capabilities
\citep{dai2023instructblip,liu2023visual,caffagni2024revolution,zhang2024mm}.
On the other hand, generative models have made breakthrough progress, enabling the creation of high-quality and high-fidelity image and video content \citep{podell2023sdxl,he2023latentvideodiffusionmodels,blattmann2023stable,gao2025seedance10exploringboundaries}.
However, diffusion-based generative models and understanding models built on the autoregressive architecture have developed independently along separate paths, making unified integration challenging \citep{zhao2025unified}.

Recently proposed unified multimodal models (UMMs) integrate understanding and generation capabilities within a single framework.
These models unify cross-modal understanding and generation within a single framework, providing a consistent interaction interface and enabling joint optimization of understanding and generation, which better supports interleaved multimodal interaction, multi-turn multimodal dialogue, and unified understanding-and-generation workflows \citep{tian2024mminterleavedinterleavedimagetextgenerative,zhan-etal-2024-anygpt,zhou2025opening,deng2025emerging,zhang2025unimodel}.
However, current UMMs still face issues such as the imbalance between understanding and generation capabilities and the lack of efficient integrated architectural designs, which limit their performance in practical tasks \citep{zhao2025unified}.

Evaluating UMMs in realistic multi-turn dialogues is an essential task for their future development.
However, existing benchmarks fail to cover this critical area and still have limitations.
First, most benchmarks adopt single-turn settings that evaluate either understanding or generation in isolation, failing to capture realistic multi-turn interleaved image--text dialogue and thus preventing unified evaluation \citep{li2025unieval,xie2025mme}.
Second, even when multi-turn interleaving is considered, some benchmarks still rely on static question answering and fail to update ground-truth answers based on the model’s responses, making it difficult to objectively assess context awareness in long-horizon dialogues \citep{chow_weave_2025}.
Third, existing benchmarks often stop at capability scoring and listing observations; although they can reveal the limitations above, practical mitigation strategies for these phenomena remain relatively scarce at present \citep{xia2024mmie,chen2024interleaved,zhou2025opening}.
\looseness=-1 Therefore, there is an urgent need to build new benchmarks to conduct in-depth, dynamic, and systematic analysis of UMMs.

\begin{figure*}[t]
    \centering
    \includegraphics[width=1\linewidth]{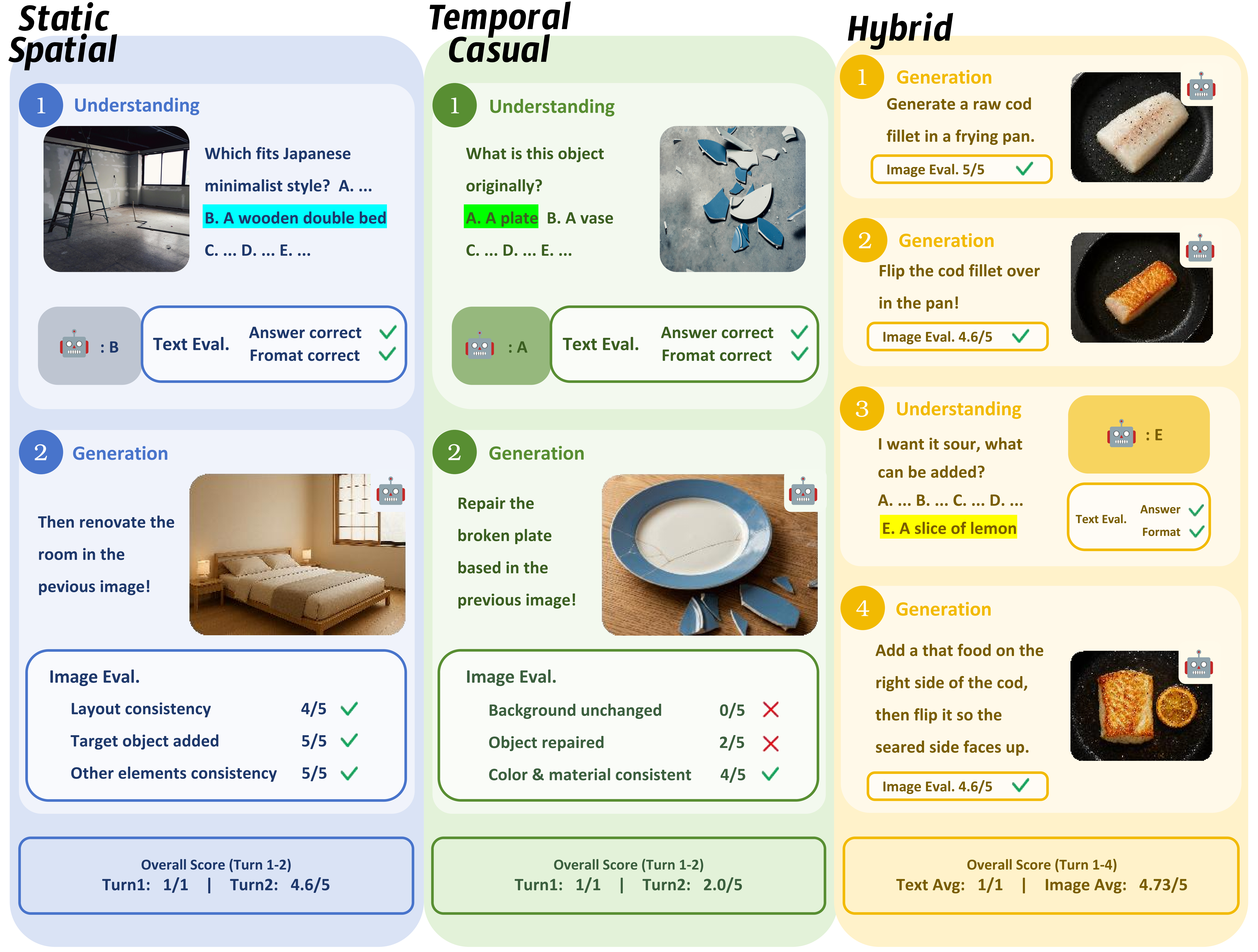}
    \caption{Overview of IMUG-Bench. IMUG-Bench evaluates UMMs’ understanding and generation capabilities through multi-turn interleaved interactions.}
    \label{fig:imug_overview}
\end{figure*}


To bridge these gaps, we propose IMUG-Bench, a benchmark for \textbf{I}nterleaved \textbf{M}ultimodal \textbf{U}nderstanding and \textbf{G}eneration.
Fig.~\ref{fig:imug_overview} illustrates our benchmark.
It is a comprehensive benchmark specifically designed to jointly evaluate the understanding and generation capabilities of UMMs in multi-turn interleaved interactions.
We first manually construct templates for questions, answers, and evaluation-points, and then call an LLM to generate keyword sequences to populate the templates, after which we manually filter these sequences and collect multi-source images, and finally use an LLM or VLM to fill the templates.
IMUG-Bench comprises three classes: Static Spatial, Temporal Causal, and Hybrid, and covers 19 domains and 97 tasks, totaling 3,113 samples and 12,034 dialogue turns.
Overall, IMUG-Bench provides a unified evaluation of UMMs across multi-turn interleaved settings.
IMUG-Bench differs from the other benchmarks, as shown in Table~\ref{tab:benchmark_comparison_interaction}.

Using IMUG-Bench, we systematically evaluate current mainstream open-source and closed-source UMMs.
The results show that current mainstream models still fall short on long-context, multi-turn interleaved generation.
As the content length and number of interaction turns increase, models exhibit pronounced exposure bias in generation tasks, and even top closed-source models still have substantial room for improvement.
In addition, experiments show that introducing test-time scaling with Chain-of-Thought, Self-Verification, and Best-of-$N$ Sampling can improve performance and mitigate exposure bias, offering new insights into enhancing the robustness of UMMs.

In summary, the main contributions of this work are as follows:
\begin{itemize}[leftmargin=2em]
    \item \textbf{IMUG-Bench benchmark construction}: We propose IMUG-Bench, a multi-turn interleaved benchmark with broad domain coverage, comprising 3,113 samples and 12,034 dialogue turns, enabling fine-grained assessment of UMMs' understanding and generation capabilities.
    \item \textbf{Systematic evaluations and analysis}: Through extensive experiments, we systematically evaluate mainstream open-source and closed-source UMMs, characterize their capability boundaries in multi-turn interleaved interactions, and identify exposure bias in generation tasks.
    \item \textbf{Optimization with test-time scaling}: We demonstrate using test-time scaling methods including Chain-of-Thought, Self-Verification, and Best-of-$N$ Sampling mitigates exposure bias, providing insights into enhancing UMMs' multi-turn interaction capability.

\end{itemize}

\begin{table}[t]
  \caption{Comparison of unified multimodal benchmarks. Abbreviations: Modal. = Modality; Mult. Iter. = Multi-turn Interaction; Dyna. Und. = Dynamic Understanding; Expos. = Exposure.}
  \label{tab:benchmark_comparison_interaction}
  \centering
  \setlength{\tabcolsep}{4.5pt} 
  \renewcommand{\arraystretch}{1.02} 

  \begin{threeparttable}
  \begin{tabular*}{\linewidth}{@{\extracolsep{\fill}} l @{} c c c c c c c}
    \toprule
    \textbf{Benchmark} &
    \textbf{Samples} &
    \makecell[c]{\textbf{Iter.}\\\textbf{Turns}} &
    \makecell[c]{\textbf{Modal.}\\\textbf{Unified}} &
    \makecell[c]{\textbf{Mult.}\\\textbf{Iter.}} &
    \makecell[c]{\textbf{Dyna.}\\\textbf{Und.}} &
    \makecell[c]{\textbf{Expos.}\\\textbf{Bias}} &
    \makecell[c]{\textbf{Test-time}\\\textbf{Scaling}} \\
    \midrule
    CoMM\tnote{1}             & 500    & 500    & \cmark & \xmark & \xmark & \xmark & \xmark \\
    UniEval\tnote{2}          & 4,231  & 4,321  & \xmark & \xmark & \cmark & \xmark & \xmark \\
    MMIE\tnote{3}             & 20,103 & 20,103 & \cmark & \xmark & \xmark & \xmark & \xmark \\
    OpenING\tnote{4}          & 5,400  & 5,400  & \cmark & \xmark & \xmark & \xmark & \xmark \\
    ISG-Bench\tnote{5}        & 1,150  & 1,150  & \cmark & \xmark & \xmark & \xmark & \xmark \\
    MME-U\tnote{6}            & 4,104  & 4,104  & \xmark & \xmark & \xmark & \xmark & \xmark \\
    Uni-MMMU\tnote{7}         & 885    & 885    & \cmark & \xmark & \cmark & \xmark & \xmark \\
    WEAVE Bench\tnote{8}      & 100    & 367    & \cmark & \cmark & \xmark & \xmark & \xmark \\
    \midrule
    IMUG-Bench                & 3,113  & 12,034 & \cmark & \cmark & \cmark & \cmark & \cmark \\
    \bottomrule
  \end{tabular*}

  \begin{tablenotes}[flushleft]
    \footnotesize
    \item[] [1]~\cite{chen2025comm}; [2]~\cite{li2025unieval}; [3]~\cite{xia2024mmie}; [4]~\cite{zhou2025opening}; [5]~\cite{chen2024interleaved}; [6]~\cite{xie2025mme}; [7]~\cite{zou2025uni}; [8]~\cite{chow_weave_2025}.
  \end{tablenotes}
  \end{threeparttable}
\end{table}

\section{Related Work}
\subsection{Unified Multimodal Models}
The rapid evolution of autoregressive LLMs has driven the development of MLLMs, which align cross-modal features into the text token space for fine-grained visual grounding and robust cross-modal instruction following \citep{wu2023multimodal,wang2024cogvlm,liang2024comprehensive,yin_survey_2024}.
Meanwhile, image generation models based on diffusion or flow matching have also made significant progress, enabling controllable and high-fidelity image synthesis, and they are gradually expanding to broader tasks such as 3D and video generation \citep{lipman2022flow,zhang2023survey,chen2025comprehensive}.
With these advances, UMMs aim to integrate understanding and generation, which enables a unified interleaved multimodal interaction paradigm with mixed text and images for both input and output \citep{zhao2025unified,yang2025survey}.
This architecture facilitates cross-modal semantic sharing, thereby forming a closed loop between understanding and generation over multiple turns \citep{chen2025blip3,wu2025omnigen2,deng2025emerging,zhang2026pelicanunify10unifiedembodied}.
Early efforts achieved system-level integration by employing an LLM as a central controller to orchestrate isolated expert models \citep{shen2023hugginggpt,yang2023mm,wu2023visual,suris2023vipergpt}.
More recent work explores different model architectures to unify the modeling of multimodal understanding and generation \citep{xie2024show,li2025dual,shi2025muddit}.
However, current unified models still face persistent challenges in balancing understanding and generation, as well as transferring LLM knowledge and reasoning to generation \citep{zhao2025unified,xiao2025omnibridge,su2025unigame}.

\subsection{Benchmarks for Unified Multimodal Models}
For understanding tasks, existing benchmarks have evolved from basic visual perception toward fine-grained understanding and multimodal reasoning, and have further expanded to subject-specific problems and structured or heterogeneous visual inputs such as tables and charts \citep{yu2023mm,liu2024mmbench,yue2024mmmu}.
For generation tasks, existing benchmarks have progressed from measuring text--image alignment to evaluating controllability and fidelity, and increasingly incorporate overall quality and human preference \citep{ghosh2023geneval,liu2023character,wu2023human}.
With the rise of unified models, unified multimodal benchmarks are also being released and improved step by step.
These benchmarks evaluate aspects such as separate evaluation of understanding and generation \citep{xie2025mme,li2025unieval}, cross-modal consistency and alignment of outputs \citep{chen2024interleaved,zhou2025opening}, and the mutual enhancement between understanding and generation \citep{zou2025uni}.
However, dynamic interaction in long-horizon scenarios is a crucial capability for UMMs in real-world applications.
Current unified multimodal benchmarks are often limited to relatively static settings and fail to evaluate this important task \citep{chow_weave_2025}.
To fill this critical gap, we propose IMUG-Bench, which uses multi-turn and mixed text--image dialogue to conduct a detailed joint evaluation of UMMs' understanding and generation capabilities.

\section{Methodology}
\subsection{Preliminary}
IMUG-Bench is a comprehensive benchmark for UMMs with multi-turn, interleaved, and dynamic interactions, designed to faithfully capture the challenges UMMs face in real-world applications.
We divide the content of IMUG-Bench into three classes: Static Spatial, Temporal Causal, and Hybrid \citep{huang2025spatial,du2025svlta,xie2025gtr,cheng2025v}.
Specifically, tasks in the Static Spatial class focus on evaluating the model's ability to recognize and generate target-element attributes, as well as ownership and spatial relationships among elements.  
Tasks in the Temporal Causal class emphasize the model's ability to reason about implicit natural laws or common-sense causal rules in the given context.  
The Hybrid class covers both classes above and uses a larger and finer set of domains to reconstruct common daily scenarios.
We broadly collect and summarize daily scenarios that match the characteristics of these three classes, and keep those that support multi-turn interaction with UMMs.
We extract the representative question directions in each scenario as domains under each class, and abstract more detailed multiple tasks.
\looseness=-1 This results in a multi-level structure for IMUG-Bench, ensuring broad coverage of common domains and specific scenarios.

\subsection{Benchmark Construction}
\begin{figure}[tb]
    \centering
    \includegraphics[width=1.0\linewidth]{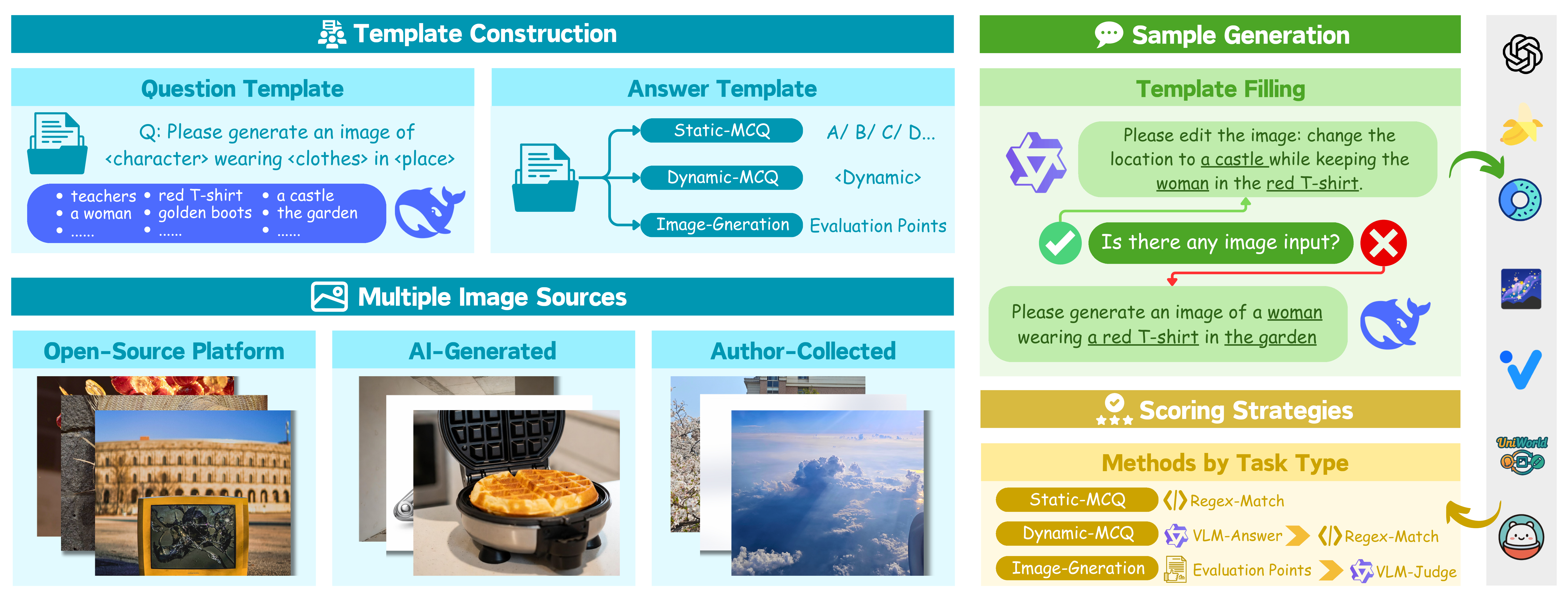}
    \caption{The IMUG-Bench construction and evaluation pipeline: manually designed templates are populated via LLM/VLM filling using images from multiple sources, and samples are scored with task-specific strategies.}
    \label{fig:construction}
\end{figure}

\looseness=-1 In IMUG-Bench, all questions are constructed via a pipeline of manual template design, LLM/VLM-based filling, and two-person human verification at every stage of LLM/VLM outputs.
The detailed benchmark construction pipeline is illustrated in Figure~\ref{fig:construction}. 
Detailed statistics are provided in Section~\ref{statistics}.

\textbf{\textit{Template Design.}} Each task instance in IMUG-Bench is a multi-turn interaction, and each turn belongs to one of three question types: Static-MCQ and Dynamic-MCQ for evaluating understanding, and Image-Generation for evaluating generation.
Specifically:
\begin{itemize}[leftmargin=2em]
    \item \textbf{Static-MCQ} is a multiple-choice question with fixed correct answers. It is used to evaluate the model's understanding ability under stable multimodal context.
    \item \textbf{Dynamic-MCQ} is a multiple-choice question whose correct answer depends on an image generated by the model in previous turns. It is used to evaluate the model's ability to understand its own generated images during multi-turn interaction. It is also introduced to address the limitation that existing benchmarks cannot measure such ability in realistic settings.
    \item \textbf{Image-Generation} requires the model to generate an image according to the instruction, optionally with reference images. It is used to evaluate the model's generation ability, including instruction following, visual consistency, and multi-turn editing performance.
\end{itemize}

For these question types, we manually design question templates for each task in IMUG-Bench to match the corresponding understanding or generation content.
\looseness=-1 In addition, we determine an appropriate number of turns and the input--output modality configuration for each turn. More statistics of IMUG-Bench, including its domain/task distribution, turn counts, and interleaved modality sequences, are provided in Appendix~\ref{sec:appendix_data_overview}.

Meanwhile, for every turn except the first, we annotate its dependencies on previous turns by specifying the indices of the turns that must be referenced.
Finally, we leave blanks for key words that determine the main content of each question.
Based on the content directions of each template, we use an LLM to enumerate fillable keyword key--value pairs.

Furthermore, we design answer templates for the three question types.
For Static-MCQ, we specify the option construction logic in the template, and generate the options with LLM/VLM during question filling.
For Dynamic-MCQ, we design the option section as a \texttt{<DYNAMIC>} placeholder.
For Image-Generation, we provide a fixed set of evaluation-points (both in number and evaluation aspects), where keywords are left as placeholders to be filled.
Finally, we conduct a two-person manual check of images, question templates, keyword lists, and answer templates, removing or fixing problematic samples.

\textbf{\textit{Samples Generation.}} After the template design and verification, we use LLMs/VLMs to appropriately fill the templates and form the final samples.
Specifically, we diversify the template filling based on the keyword lists to cover as many instantiated questions as possible.
We use an LLM to fill templates for text-only questions, and a VLM to fill templates for questions with image inputs.

During sample generation, we populate the pre-defined answer templates for the three question types separately.
For Static-MCQ, an LLM or a VLM directly fills the templates to generate samples.
For Dynamic-MCQ, we use a VLM to determine the ground-truth answer based on the prompt image, the prompt text, and the options only after the model has finished answering.
For Image-Generation, we fill the keywords into the evaluation-points template.
Beyond keyword filling, the LLM and VLM are also instructed to smooth the question wording during generation, avoiding mechanical keyword insertion, while keeping the original meaning unchanged and preventing any content leakage.
Before large-scale template filling, we conduct a small-scale manual verification to check the quality of the constructed samples.
After large-scale construction, we further conduct a two-person manual review of all task samples to ensure that the final benchmark does not contain factual errors.
Details of the image sources and curation process are provided in Appendix~\ref{sec:appendix_image_sources}. Detailed prompt templates used for question filling in the benchmark construction process are provided in Appendix~\ref{sec:appendix_prompts_question_filling}.

\enlargethispage{0.7\baselineskip}

\subsection{IMUG-Bench Evaluation}
\label{evaluation}
Each turn in IMUG-Bench can be categorized into text questions and image questions according to the output modality.
For questions with different output modalities, we design separate evaluation and scoring strategies to more properly quantify model performance.

\textit{\textbf{Understanding.}} For the understanding tasks in Static-MCQ and Dynamic-MCQ, the model must determine both the correct options and the number of correct options, and output them in the required format.
We then apply a regex-based rule scoring scheme.
$$
S_{\mathrm{MCQ}}
= w_{\mathrm{fmt}} \,\bigl(n_{\mathrm{corr}}-n_{\mathrm{incorr}}\bigr)\,u  .
$$
Here,  \(w_{\mathrm{fmt}}\) is the formatting weight; \(n_{\mathrm{corr}}\) and \(n_{\mathrm{incorr}}\) are the numbers of correct and incorrect options selected, respectively, and \(u\) is the unit score per correct option, defined as \(1\) divided by the number of correct options.

The formatting weight is defined as follows: (A) $w_{\mathrm{fmt}}{=}1$ if the output is a continuous sequence of uppercase option letters with no separators; (B) $w_{\mathrm{fmt}}{=}0.75$ if the output contains uppercase option letters but deviates from the required format (e.g., separated by symbols or accompanied by the option text), while containing no additional explanatory text; (C) $w_{\mathrm{fmt}}{=}0.5$ if the output includes uppercase option letters together with extra explanatory text beyond the selected options. (D) If no valid option letter is output, we set $w_{\mathrm{fmt}}{=}0$.

\textit{\textbf{Generation.}} Due to the open-ended nature of image generation outputs, it is often difficult for evaluators to score image outputs with rule-based methods.
Therefore, following the VLM-as-a-Judge paradigm, we adopt dynamic evaluation-points filling to score tasks whose output modality is image, and we further study the validity of the scoring results via correlation analysis.
We further validate the reliability of this scoring protocol against two-person human ratings; the results are reported in Appendix~\ref{app:human_validation}.

Specifically, we decompose the evaluation of a single image into VLM-based scoring over multiple personalized evaluation-points. 
An evaluation point represents an independent evaluation dimension that measures how well the generated image satisfies a specific aspect of the user requirement, and different evaluation-points are complementary to each other.
The evaluation-points are grouped into two types by content:
(1) \emph{Image update requirements}, which specify the expected content for the output (or the intended changes when reference images are available) and check whether the instruction is correctly followed, including targeted generation and style adjustment;
(2) \emph{Image consistency checks}, which require the output image to preserve element attributes that should remain unchanged based on the reference images from previous turns; this type applies only when such references exist.
Each image question is assigned multiple evaluation-points according to its target skills, ensuring coverage of key factors that affect image quality and requirement fulfillment.

During scoring, each output image is scored independently, and when needed, the VLM judge is provided with previous images as references to evaluate consistency fulfillment.
The VLM judge assigns an integer score from 0 to 5 (inclusive) to each evaluation-point: higher scores indicate better fulfillment and alignment with the evaluation-point; 0 indicates no relevant output or a completely wrong result, while 5 indicates perfect compliance with no errors.
In each turn, the total image score is computed as:
$$
S_{\mathrm{img}}=\frac{\sum_{i=1}^{N} s_i}{N \cdot 5}
$$
Here, $N$ denotes the number of evaluation-points, and $s_i$ is the score of the $i$-th point.

\subsection{Benchmark Statistics}
\label{statistics}
IMUG-Bench is organized into three classes by task content: Static Spatial, Temporal Causal, and Hybrid.
Within these classes, we define 19 domains based on application contexts and knowledge areas, and further divide them into 97 tasks by more fine-grained question content and template structures, resulting in 3{,}113 task samples in total.

Each task sample consists of a multi-turn interaction with either text-only or text-and-image inputs, and requires the model to generate outputs in an interleaved text--image format.
The number of turns ranges from 2 to 6, with 12{,}034 turns in total and an average of 3.87 turns per task.
The counts of understanding and generation turns are relatively balanced, which helps ensure broad and balanced evaluation coverage. Pie-chart visualizations of the class/domain distribution and a word-cloud summary of high-frequency concepts are provided in Appendix~\ref{sec:appendix_data_overview}, Fig.~\ref{fig:world_cloud}. More fine-grained statistics, including the full domain--task breakdown, turn counts, and modality sequences, are provided in Appendix~\ref{sec:appendix_data_overview}, Table~\ref{tab:dataset_stats}.


\subsection{Test-time Scaling} 
\label{testtime}
Our experiments reveal exposure bias in image questions: as the number of turns increases, UMM performance degrades.
We then explore three test-time scaling methods to improve the models and mitigate the exposure bias caused by multi-turn dialogue.

\textit{\textbf{Chain-of-Thought.}} Given the strong effectiveness of Chain-of-Thought in language-model reasoning \cite{wei2022chain}, we introduce it to enhance instruction alignment over multi-turn context.
Specifically, before answering image questions, we require the model to summarize the current image-generation request and explicitly list its generation plan in the thinking stage, and then generate the image accordingly.
This reasoning content is not appended to the dialogue history.

\textit{\textbf{Self-Verification.}} Since UMMs typically have stronger understanding than generation, we introduce a closed-loop Self-Verification mechanism that uses the model's understanding capability to check instruction following and cross-turn consistency of generated images, and uses the diagnostic feedback to iteratively redraw and repair.
At turn $t$, the model first generates an image, denoted as $R_t$.
The model is then asked to assess whether $R_t$ satisfies the current instruction and preserves the required style and visual details from the previous context.
If the model judges that $R_t$ is acceptable, this image is directly used as the output of the current turn.
Otherwise, the model provides review feedback that identifies the main problems and suggests how the image should be improved.
This feedback is then combined with the dialogue context and used to guide a new round of image generation, producing an updated image for the same turn. To keep the method practical, we set a maximum number of verification rounds.
If no satisfactory result is obtained within this limit, the latest generated image is kept as the final output for the current turn.
Only the final accepted image is added to the dialogue history for later turns, while intermediate failed images and temporary review feedback are discarded after the current turn.

\textit{\textbf{Best-of-$N$ Sampling.}} Since image generation is stochastic, for an image question we generate multiple images under the same history and request, then use the model's understanding ability to select and keep the best match, increasing the chance of a high-quality result.
(1) Given the same history and the current user request, the model generates $N$ candidate images $R_{t,1}, R_{t,2}, \ldots, R_{t,N}$.
(2) The system caches all $N$ candidates, provides them together with the dialogue history, and prompts the model to select the one that best fulfills the current request and necessary consistency.
(3) The model evaluates each candidate's fulfillment and outputs the index of the selected best candidate as the turn result, then proceeds to the next turn.
In this mechanism, unselected candidate images and the explicit fulfillment evaluations are not included in the history for subsequent turns; only the selected best candidate is kept as the official output and appended to the history.

\begin{table}[tb]
  \caption{Main results on IMUG-Bench. We report understanding (Und.) and generation (Gen.) scores across the three classes (Static Spatial, Temporal Causal, Hybrid), along with the overall average (Avg); the top-3 results in each metric are highlighted as \textcolor{medalgold}{\underline{\textbf{gold}}} (best), \textcolor{medalsilver}{\underline{\textbf{silver}}} (second best), and \textcolor{medalbronze}{\underline{\textbf{bronze}}} (third best).}
  \label{tab:model_cost}
  \centering
  
  \newcolumntype{M}[1]{>{\centering\arraybackslash}m{#1}}
  \newcolumntype{Y}{>{\centering\arraybackslash}X}
  \newcolumntype{L}[1]{>{\raggedright\arraybackslash}m{#1}} 

  \begin{threeparttable}
  \begin{tabularx}{\linewidth}{@{}L{0.2\linewidth} Y Y *{8}{Y}@{}}
    \toprule
    \multirow{2}{*}{\makecell[c]{\textbf{Model Name}}} &
    \multirow{2}{*}{\makecell[c]{\textbf{Open}\\\textbf{Source}}} &
    \multirow{2}{*}{\textbf{Size}} &
    \multicolumn{2}{c}{\textbf{St.\&Sp.}} &
    \multicolumn{2}{c}{\textbf{Te.\&Ca.}} &
    \multicolumn{2}{c}{\textbf{Hybrid}} &
    \multirow{2}{*}{\textbf{Avg}} \\
    \cmidrule(lr){4-5}\cmidrule(lr){6-7}\cmidrule(lr){8-9}
    & & & \textbf{Und.} & \textbf{Gen.} & \textbf{Und.} & \textbf{Gen.} & \textbf{Und.} & \textbf{Gen.} & \\
    \midrule

    GPT-5 Image Mini\tnote{1} & \xmark & - &
    \medal{medalgold}{81.6} & \medal{medalgold}{88.0} &
    \medal{medalgold}{74.8} & \medal{medalgold}{83.1} &
    \medal{medalgold}{80.9} & \medal{medalgold}{79.3} &
    \medal{medalgold}{82.3} \\
    Nano Banana\tnote{2} & \xmark & - &
    \medal{medalsilver}{78.0} & \medal{medalsilver}{86.9} &
    \medal{medalsilver}{73.7} & \medal{medalsilver}{82.8} &
    \medal{medalsilver}{70.6} & \medal{medalsilver}{76.7} &
    \medal{medalsilver}{79.5} \\
    \midrule

    BAGEL\tnote{3} & \cmark & 15B & 49.7 & 57.9 & 53.8 & 49.3 & 36.4 & 52.8 & 51.9 \\
    BLIP3-o 4B\tnote{4} & \cmark & 10B & 53.5 & 62.4 & 62.1 & 59.3 & \medal{medalbronze}{55.8} & 60.6 & 59.3 \\
    BLIP3-o 8B\tnote{4} & \cmark & 14B & 65.8 & \medal{medalbronze}{67.3} & 68.9 & \medal{medalbronze}{61.1} & 52.4 & \medal{medalbronze}{67.8} & \medal{medalbronze}{65.0} \\
    OmniGen2\tnote{5} & \cmark & 7B  & \medal{medalbronze}{67.7} & 44.4 & \medal{medalbronze}{70.0} & 43.4 & 55.2 & 50.2 & 55.1 \\
    Ovis-U1\tnote{6} & \cmark & 3B  & 54.0 & 45.7 & 51.6 & 49.3 & 48.5 & 54.2 & 50.0 \\
    UNIWORLD-V1\tnote{7} & \cmark & 20B & 29.3 & 63.9 & 28.4 & 55.5 & 24.2 & 57.8 & 46.6 \\

    \bottomrule
  \end{tabularx}

\begin{tablenotes}[flushleft]
  \footnotesize
  \item[] [1]~\cite{singh2025openai}; 
         [2]~\cite{comanici2025gemini}; 
         [3]~\cite{deng2025emerging}; 
         [4]~\cite{chen2025blip3}; 
         [5]~\cite{wu2025omnigen2}; 
         [6]~\cite{wang2025ovis}; 
         [7]~\cite{lin2025uniworld}.
\end{tablenotes}
  \end{threeparttable}
\end{table}

\section{Experiment}
\subsection{Experimental Setup}
We evaluate eight UMMs, including six open-source UMMs (BAGEL~\cite{deng2025emerging}, BLIP3-o 4B~\cite{chen2025blip3}, BLIP3-o 8B~\cite{chen2025blip3}, OmniGen2~\cite{wu2025omnigen2}, Ovis-U1~\cite{wang2025ovis}, and UNIWORLD-V1~\cite{lin2025uniworld}) and two closed-source UMMs (Nano Banana (Gemini 2.5 Flash Image)~\cite{comanici2025gemini} and GPT-5 Image Mini (GPT-5 Mini + GPT Image 1 Mini)~\cite{singh2025openai}).

\textbf{History Construction.}
For open-source models, we concatenate the user queries and model answers in order, add the necessary role tags, and feed the full history back to the model.
BAGEL supports interleaved text--image inputs in the original order, so we rebuild the history in the exact interaction order and return it to the model.
In contrast, BLIP3-o, OmniGen2, Ovis-U1, and UNIWORLD-V1 only support a serialized image-list input, so we represent the history as an image list with in-text placeholders.
In particular, since Ovis-U1 does not support multi-image editing, we horizontally concatenate all historical input and output images, assign indices, and add a brief prompt that specifies the image-input rules during the dialogue.
For closed-source models, we use their official multi-turn dialogue APIs to run multi-turn conversations with history memory.

\textbf{Scoring Inputs and Model Selection.}
We follow Section~\ref{evaluation} for scoring.
For Dynamic-MCQ, the judge takes as input the question and options of the target turn, together with the full context of the annotated history-related turns, and returns the set of ground-truth answer options.
For Image-Generation turns, the judge takes as input the question of the target turn, the evaluation-point list, the model-generated image, and the annotated history-related images, and returns structured per-point scores.
For benchmark construction, we use DeepSeek-V3.2 and Qwen3-VL-235B-A22B-Thinking. For scoring, we use Qwen3-VL-235B-A22B-Thinking as the judge. The prompts used for model evaluation, dynamic MCQ answer retrieval, and image assessment are provided in Appendix~\ref{sec:appendix_prompts_evaluation}--\ref{sec:appendix_prompts_image_assessment}.

\subsection{Main Results on IMUG-Bench}
\label{main_result}

\begin{table}[tb]
  \caption{Per-turn results of all evaluated models on IMUG-Bench. We report scores separately for text and image questions from Turn 1 to Turn 6, along with the overall average (Avg). Mod. denotes the question modality. The top-3 overall average results are highlighted as \textcolor{medalgold}{\underline{\textbf{gold}}} (best), \textcolor{medalsilver}{\underline{\textbf{silver}}} (second best), and \textcolor{medalbronze}{\underline{\textbf{bronze}}} (third best).}
  \label{tab:turn_wise_scores}
  \centering
  
  \newcolumntype{M}[1]{>{\centering\arraybackslash}m{#1}}
  \newcolumntype{L}[1]{>{\raggedright\arraybackslash}m{#1}} 

  \begin{threeparttable}
  \fontsize{9.3}{8}\selectfont
  \setlength{\tabcolsep}{3.2pt}
  \renewcommand{\arraystretch}{1.05}
  \begin{tabular*}{\linewidth}{@{\extracolsep{\fill}} M{0.19\linewidth} M{0.06\linewidth} M{0.055\linewidth} M{0.06\linewidth} *{6}{M{0.055\linewidth}} M{0.055\linewidth}@{}}
    \toprule
    \multirow{2}{*}[-0.45ex]{\makebox[0.19\linewidth][c]{\makecell[c]{\textbf{Model Name}}}} &
    \multirow{2}{*}[-0.45ex]{\makebox[0.06\linewidth][c]{\makecell[c]{\textbf{Open}\\\textbf{Source}}}} &
    \multirow{2}{*}[-0.45ex]{\makebox[0.055\linewidth][c]{\makecell[c]{\textbf{Size}}}} &
    \multirow{2}{*}[-0.45ex]{\makebox[0.06\linewidth][c]{\makecell[c]{\textbf{Mod.}}}} &
    \multicolumn{6}{c}{\textbf{Turn}} &
    \multirow{2}{*}[-0.45ex]{\makebox[0.055\linewidth][c]{\makecell[c]{\textbf{Avg}}}} \\
    \cmidrule(lr){5-10}
    & & & & \textbf{T1} & \textbf{T2} & \textbf{T3} & \textbf{T4} & \textbf{T5} & \textbf{T6} & \\
    \midrule

        \multirow{2}{*}[-0.51ex]{GPT-5 Image Mini\tnote{1}} & \multirow{2}{*}[-0.51ex]{\xmark} & \multirow{2}{*}[-0.51ex]{-} & Text  &
    73.8 & 84.8 & 86.1 & 76.0 & 71.2 & 77.2 & \multirow{2}{*}[-0.51ex]{\medal{medalgold}{82.3}} \\
    \cmidrule(lr){4-10}
    & & & Image &
    94.3 & 84.9 & 79.1 & 78.1 & 77.5 & 54.8 & \\
    \midrule

    \multirow{2}{*}[-0.51ex]{Nano Banana\tnote{2}} & \multirow{2}{*}[-0.51ex]{\xmark} & \multirow{2}{*}[-0.51ex]{-} & Text  &
    69.7 & 81.7 & 81.7 & 73.0 & 65.2 & 72.8 & \multirow{2}{*}[-0.51ex]{\medal{medalsilver}{79.5}} \\
    \cmidrule(lr){4-10}
    & & & Image &
    92.9 & 82.3 & 74.7 & 73.2 & 72.7 & 58.7 & \\
    \midrule

    \multirow{2}{*}[-0.51ex]{BAGEL\tnote{3}} & \multirow{2}{*}[-0.51ex]{\cmark} & \multirow{2}{*}[-0.51ex]{15B} & Text  &
    47.8 & 49.9 & 48.8 & 50.7 & 46.4 & 44.8 & \multirow{2}{*}[-0.51ex]{51.9} \\
    \cmidrule(lr){4-10}
    & & & Image &
    68.5 & 52.4 & 47.5 & 43.2 & 42.7 & 24.7 & \\
    \midrule

    \multirow{2}{*}[-0.51ex]{BLIP3-o 4B\tnote{4}} & \multirow{2}{*}[-0.51ex]{\cmark} & \multirow{2}{*}[-0.51ex]{10B} & Text  &
    52.3 & 60.6 & 62.7 & 58.2 & 42.8 & 66.9 & \multirow{2}{*}[-0.51ex]{59.3} \\
    \cmidrule(lr){4-10}
    & & & Image &
    77.7 & 57.0 & 53.0 & 51.0 & 50.0 & 48.9 & \\
    \midrule

    \multirow{2}{*}[-0.51ex]{BLIP3-o 8B\tnote{4}} & \multirow{2}{*}[-0.51ex]{\cmark} & \multirow{2}{*}[-0.51ex]{14B} & Text  &
    56.4 & 69.9 & 66.0 & 63.1 & 62.4 & 74.4 & \multirow{2}{*}[-0.51ex]{\medal{medalbronze}{65.0}} \\
    \cmidrule(lr){4-10}
    & & & Image &
    80.1 & 66.0 & 56.1 & 53.0 & 52.9 & 51.7 & \\
    \midrule

    \multirow{2}{*}[-0.51ex]{OmniGen2\tnote{5}} & \multirow{2}{*}[-0.51ex]{\cmark} & \multirow{2}{*}[-0.51ex]{7B} & Text  &
    71.7 & 67.7 & 64.4 & 63.8 & 61.0 & 65.1 & \multirow{2}{*}[-0.51ex]{55.1} \\
    \cmidrule(lr){4-10}
    & & & Image &
    54.0 & 43.3 & 39.0 & 38.6 & 35.3 & 31.4 & \\
    \midrule

    \multirow{2}{*}[-0.51ex]{Ovis-U1\tnote{6}} & \multirow{2}{*}[-0.51ex]{\cmark} & \multirow{2}{*}[-0.51ex]{3B} & Text  &
    49.4 & 51.6 & 50.5 & 54.3 & 53.2 & 58.5 & \multirow{2}{*}[-0.51ex]{50.0} \\
    \cmidrule(lr){4-10}
    & & & Image &
    61.8 & 47.1 & 41.1 & 40.9 & 36.9 & 27.3 & \\
    \midrule

    \multirow{2}{*}[-0.51ex]{UNIWORLD-V1\tnote{7}} & \multirow{2}{*}[-0.51ex]{\cmark} & \multirow{2}{*}[-0.51ex]{20B} & Text  &
    21.9 & 28.8 & 29.1 & 32.1 & 28.6 & 27.6 & \multirow{2}{*}[-0.51ex]{46.6} \\
    \cmidrule(lr){4-10}
    & & & Image &
    78.3 & 60.1 & 49.7 & 44.5 & 43.9 & 42.9 & \\
    \bottomrule
  \end{tabular*}

\begin{tablenotes}[flushleft]
  \footnotesize
  \item[] [1]~\cite{singh2025openai}; 
         [2]~\cite{comanici2025gemini}; 
         [3]~\cite{deng2025emerging}; 
         [4]~\cite{chen2025blip3}; 
         [5]~\cite{wu2025omnigen2}; 
         [6]~\cite{wang2025ovis}; 
         [7]~\cite{lin2025uniworld}.
\end{tablenotes}
  \end{threeparttable}
\end{table}

Table~\ref{tab:model_cost} reports the understanding (Und.) and generation (Gen.) scores of each evaluated model on the three IMUG-Bench classes (Static Spatial, Temporal Causal, and Hybrid), together with the overall average score (Avg).
In the overall ranking, closed-source models lead by a clear margin: GPT-5 Image Mini achieves the highest understanding and generation scores across all three classes, followed closely by Nano Banana, and both maintain consistently strong performance across classes.
\looseness=-1 Among open-source models, BLIP3-o 8B performs best overall and ranks first among open-source models in generation, indicating relatively stable generation performance across task categories.

These results suggest the following.
First, the advantage of closed-source models is not only higher average scores but also stronger stability: their understanding and generation scores remain high across all three classes, indicating more balanced capabilities and lower sensitivity to task-distribution shifts.
In contrast, open-source models score lower overall and fluctuate more, reflecting a remaining gap in robustness.
Second, many open-source models exhibit a clear imbalance between understanding and generation.
Under IMUG-Bench's multi-turn interleaved setting, this imbalance directly hurts overall performance, because generation often depends on prior understanding and constraint alignment, and weaknesses in one side can propagate across turns and reduce the average score.
Third, all open-source models obtain low scores on four domains---Stack Simulation, Geometric Coloring, Mirror Reasoning, and Paper Crafting---indicating a shared weakness in fine-grained spatial or geometric multimodal tasks under multi-turn interleaving.
Additional radar-chart visualizations and detailed cross-domain analyses of all evaluated models are provided in Appendix~\ref{sec:appendix_radar_results}, Figs.~\ref{fig:c1}--\ref{fig:c3}.

We further compare the performance differences between Static-MCQ and Dynamic-MCQ across all evaluated models (Appendix~\ref{sec:appendix_sup_of_mcq}, Table~\ref{tab:static_dynamic_mcq}). We find that all models score substantially lower on Dynamic-MCQ than on Static-MCQ. This result indicates that, in complex interleaved multi-turn interactions, models' understanding performance is more easily affected by accumulated errors in the dialogue history, making it difficult for them to accurately understand the specific content of their own previously generated images. This phenomenon further suggests that multi-turn understanding of self-generated content remains an important challenge for current UMMs.

\subsection{Exposure Bias of UMMs}

Overall, text-question scores do not show a consistent monotonic trend as the number of turns increases; in contrast, image-question scores show a clear downward trend with more turns for almost all models.
Table~\ref{tab:turn_wise_scores} presents the turn-wise score distributions of all evaluated models on text questions and image questions. A more intuitive visualization of these trends is provided in Appendix~\ref{sec:appendix_turn_trends}, Fig.~\ref{fig:exposurebias}.

This phenomenon reveals the exposure bias of current UMMs in multi-turn interaction: unlike single-turn evaluation, multi-turn image generation must satisfy both the current instruction and cross-turn consistency constraints. Generation errors in early turns are written into later context and repeatedly exposed to the model, forcing later turns to generate conditioned on an imperfect history and leading to error accumulation and distribution drift.
Unlike single-turn evaluation, multi-turn image generation must satisfy both the current instruction and cross-turn consistency constraints.
As a result, as the number of interaction turns grows, generation scores are more prone to decay, highlighting the necessity to evaluate and mitigate exposure bias in long-horizon interleaved tasks.

\subsection{Case Study}

\begin{figure}[tb]
    \centering
    \includegraphics[width=1\linewidth]{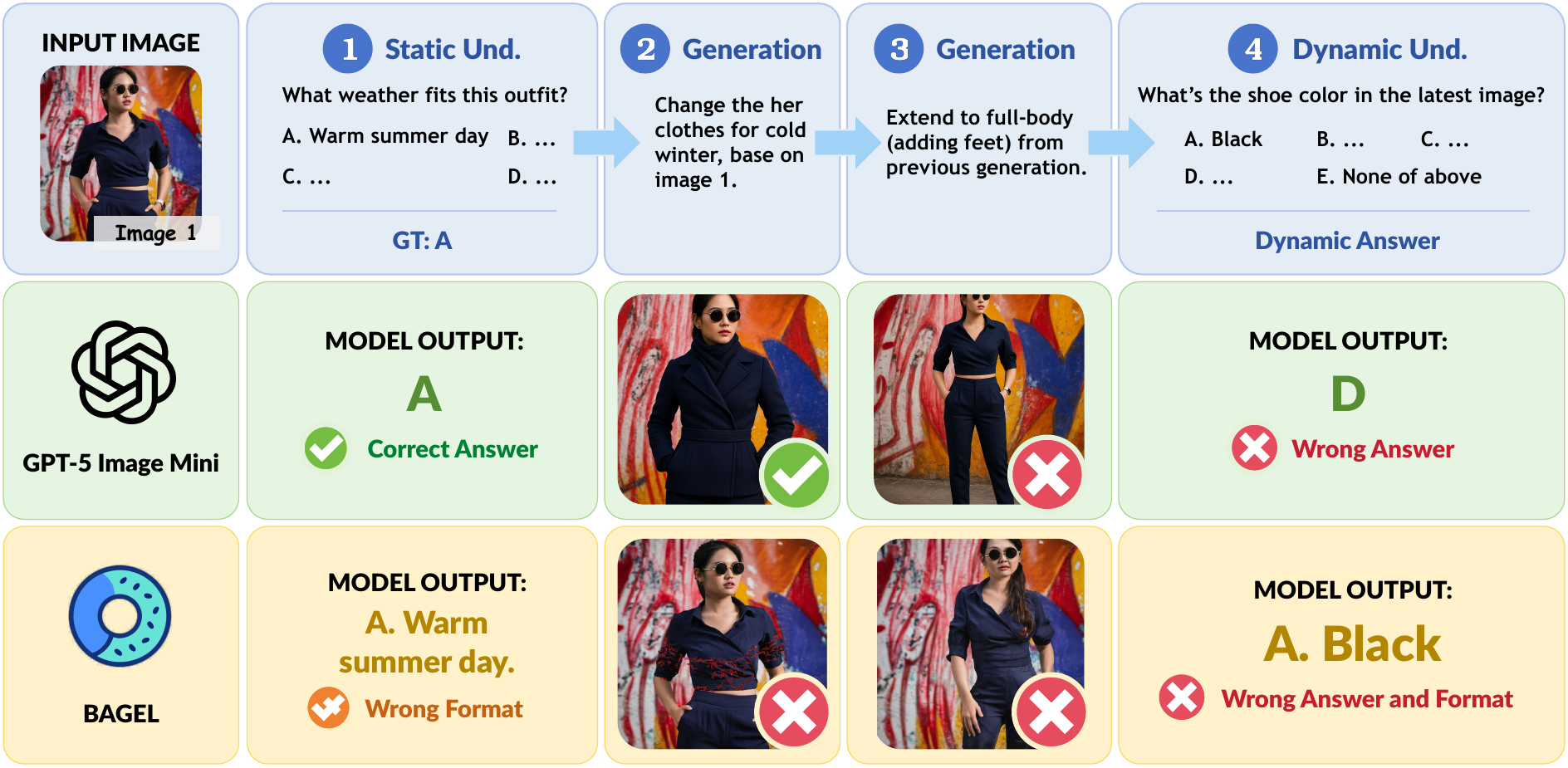}
    \caption{A representative IMUG-Bench sample illustrating the limitations of current UMMs in multi-turn interleaved interaction. Current UMMs still exhibit weak instruction-following ability, error accumulation across turns, and hallucination.}
    \label{fig:case}
\end{figure}

As shown in Fig.~\ref{fig:case}, three typical problems observed in our evaluation are illustrated by the example above.
First, closed-source models are more likely to succeed in early turns, whereas open-source models often make mistakes in output formatting and instruction following.
In this case, the closed-source models can produce the answer in the required format and complete the winter-clothing edit with higher fidelity, while the open-source models often violate the required format or miss key editing instructions, which indicates weaker instruction-following ability.

Second, for multi-turn image generation, both open-source and closed-source models show clear limitations.
In this example, although GPT-5 Image Mini successfully follows the instruction in the first image-generation turn, it fails to preserve the required edits in the next turn.
By contrast, BAGEL fails to fully satisfy the instruction in the first generation turn and then carries this error into the next turn.
This suggests that both groups of models still need improvement on generation tasks in multi-turn interaction.

Third, the final Dynamic-MCQ turn also reveals a hallucination problem.
Although neither model fully completes the previous full-body extension instruction, both of them incorrectly answer with a shoe color that does not appear in the generated image, producing a hallucinated response.


\subsection{Experiments on Test-time Scaling}
\subsubsection{Main Results}
To mitigate exposure bias in image-generation tasks as the number of turns increases, we choose BAGEL as a representative open-source UMM and re-evaluate it after applying the three test-time scaling strategies in Section~\ref{testtime} during inference. In our setup, the maximum retry count is 1, and the number of candidates $N$ for Best-of-$N$ Sampling is set to 3, which correspond to the configuration used in the right plot of Appendix~\ref{sec:appendix_turn_trends}, Fig.~\ref{fig:exposurebias}.


Results show that, compared with the original inference setting, introducing any of the strategies improves the overall image-question score and clearly slows the performance decay as turns increase.
Comparing the gains, Chain-of-Thought yields the largest improvement, suggesting that in multi-turn editing/generation, a structured summary of the current request and constraints, together with an explicit generation plan, helps reduce missed instructions and consistency violations.
Self-Verification and Best-of-$N$ Sampling also reduce error accumulation caused by low-quality outputs being written into the history, but their gains rely more on extra inference cost and the number of samples. 
Overall, these results demonstrate that test-time scaling is an effective strategy for mitigating exposure bias. 
However, the current model performance indicates that further methodological improvements are still needed to fully overcome performance decay in multi-turn generation.

\begin{table}[t]
  \caption{Image-modality scores under three test-time scaling methods and their corresponding settings. We report generation scores with the overall average score. For each metric, we first identify the best-performing setting within each test-time scaling method, and then assign medal rankings across methods based on these method-level best results; the resulting top-3 methods are highlighted as \textcolor{medalgold}{\underline{\textbf{gold}}} (best), \textcolor{medalsilver}{\underline{\textbf{silver}}} (second best), and \textcolor{medalbronze}{\underline{\textbf{bronze}}} (third best).}
  \label{tab:d0}
  \centering
  \newcolumntype{C}[1]{>{\centering\arraybackslash}m{#1}}

  \begin{tabularx}{\linewidth}{@{}
    C{0.28\linewidth}
    C{0.15\linewidth}
    C{0.19\linewidth}
    C{0.14\linewidth}
    C{0.14\linewidth}
  @{}}
    \toprule
    \textbf{Method / Setting} & 
    \textbf{Static Spatial} & 
    \textbf{Temporal Causal} & 
    \textbf{Hybrid} & 
    \textbf{Average} \\
    \midrule

    Baseline & 57.9 & 49.3 & 52.8 & 54.1 \\
    \midrule
    Chain-of-Thought & \medal{medalgold}{72.0} & \medal{medalgold}{60.1} & \medal{medalgold}{60.4} & \medal{medalgold}{65.9} \\
    \midrule
    S-V (Max=1) & 59.0 & \medal{medalbronze}{50.7} & \medal{medalbronze}{53.3} & \medal{medalbronze}{55.2} \\
    S-V (Max=2) & \medal{medalbronze}{60.6} & 49.4 & 48.2 & 54.6 \\
    S-V (Max=3) & 58.7 & 49.2 & 48.6 & 53.7 \\ 
    \midrule
    B-of-$N$ ($N$=2) & 58.0 & 50.1 & 54.9 & 54.8 \\
    B-of-$N$ ($N$=3) & 61.6 & 50.6 & 58.6 & 57.4 \\
    B-of-$N$ ($N$=4) & 61.9 & \medal{medalsilver}{52.9} & 58.0 & 58.2 \\
    B-of-$N$ ($N$=5) & \medal{medalsilver}{62.0} & 52.7 & \medal{medalsilver}{60.1} & \medal{medalsilver}{58.6} \\

    \bottomrule
  \end{tabularx}
\end{table}

\subsubsection{Ablation Study on Test-time Scaling}
To further investigate the impact of test-time scaling methods on the generative capabilities of models, we conduct ablation studies on two categories of techniques: Self-Verification and Best-of-$N$ Sampling. Using BAGEL as the base model, we examine their respective parameters and provide a comprehensive analysis of the experimental results.

\begin{itemize}[leftmargin=2em]
    \item \textbf{Ablation Study on Self-Verification}: We set the maximum retry count in the Self-Verification process to 0 (the baseline without verification), 1, 2, and 3, respectively. The resulting image-modality scores are presented in Table~\ref{tab:d0}. Analysis of the results indicates that the peak image scores across all three classes, as well as the average image score, occur between 1 or 2 retries. However, performance degrades as the retry count increases beyond 2, occasionally underperforming the baseline. This experimental evidence demonstrates that Self-Verification effectively enhances image generation quality within a certain range; however, overly frequent retries can be counterproductive.
    \item \textbf{Ablation Study on Best-of-$N$ Sampling}: In a similar vein, we evaluate the Best-of-$N$ Sampling method with $N$ ranging from 1 to 5, where $N=1$ serves as the baseline (no selection). The image scores, detailed in Table~\ref{tab:d0}, generally follow an upward trajectory with increasing $N$. Notably, the rate of improvement slows down markedly as $N$ increases, with the scores for $N=4$ and $N=5$ showing minimal variance. This evidence confirms that the quality gains are consistent rather than coincidental, while simultaneously revealing the inherent performance boundary of the approach.

\end{itemize}

\section{Conclusion}
In this paper, we propose IMUG-Bench, a large-scale benchmark with 3,113 samples and 12,034 dialogue turns, to systematically evaluate the understanding and generation capabilities of UMMs in multi-turn interleaved text--image interactions.
We conduct systematic experiments on a range of mainstream UMMs, and draw the following conclusions:
(1) Existing UMMs remain notably limited in multi-turn interleaved tasks.
As the number of turns and context length increase, generation performance commonly shows sustained decay, reflecting clear exposure bias.
(2) Closed-source models not only lead in overall average scores, but also exhibit more balanced capabilities across classes and metrics.
In contrast, open-source models score lower overall and fluctuate more, and show pronounced weaknesses on fine-grained spatial and geometry-related tasks.
(3) We validate the effectiveness of test-time scaling: introducing Chain-of-Thought, Self-Verification, and Best-of-$N$ Sampling during inference improves multi-turn image generation quality and mitigates performance decay, with Chain-of-Thought providing the largest gains.
Overall, IMUG-Bench provides a systematic foundation for studying the capability boundaries, failure modes, and optimization strategies of UMMs in multi-turn interleaved settings, and offers insights toward building more robust and controllable UMMs.

\bibliography{benchmark}
\bibliographystyle{iclr2025_conference}

\newpage
\appendix

\begin{center}
    {\LARGE \textbf{Appendix}}
\end{center}

\vspace{1em}

The appendix provides additional details and supporting materials for IMUG-Bench, organized as follows.

Appendix~\ref{sec:appendix_image_sources} describes the image sources and curation process used in benchmark construction.

Appendix~\ref{sec:appendix_data_overview} presents the data distribution of IMUG-Bench, including visualizations of the class/domain distribution, as well as detailed information regarding domain-task composition, interleaved modality sequences, and the number of interaction turns.

Appendix~\ref{sec:appendix_additional_results} reports additional experimental results, including radar-chart visualizations of task-specific model performance, supplementary results on Static-MCQ and Dynamic-MCQ, and score trends over interaction turns.

Appendix~\ref{app:human_validation} presents the human validation results for the VLM-as-a-judge scoring protocol. 

Appendix~\ref{sec:appendix_prompts_question_filling}--\ref{sec:appendix_prompts_image_assessment} provide the prompts used for calling LLMs or VLMs during question filling, model evaluation, dynamic MCQ answer and image assessment retrieval in IMUG-Bench.


\section{Image Sources and Curation}
\label{sec:appendix_image_sources}
We construct the benchmark using three complementary image sources.
\begin{itemize}
\item \textbf{\textit{Public image platforms.}} Using the finalized question templates and keyword lists, we batch-collect images that match the target scenarios from platforms such as Pexels, Pixabay, Openverse, Flickr, Unsplash, Wikimedia, and CloudAtlas.

\item \textbf{\textit{Targeted generation with image models.}} For image concepts that are difficult to collect from public platforms, we use image models such as GPT Image, Nano Banana, and Grok to generate candidate images.

\item \textbf{\textit{Real-world capture and professional content creation.}} When public platforms and model generation cannot provide high-quality images, we supplement the dataset through real-world capture or professional tools.
\end{itemize}

These sources cover diverse scenarios and styles, helping reduce source-specific bias and improve coverage of real interactive settings.
All images are manually reviewed before inclusion, and only those that match the task requirements and satisfy quality, licensing, and safety checks are retained.

\section{Data Overview of IMUG-Bench}
\label{sec:appendix_data_overview}

\begin{figure}[t]
    \centering
    \includegraphics[width=1\linewidth]{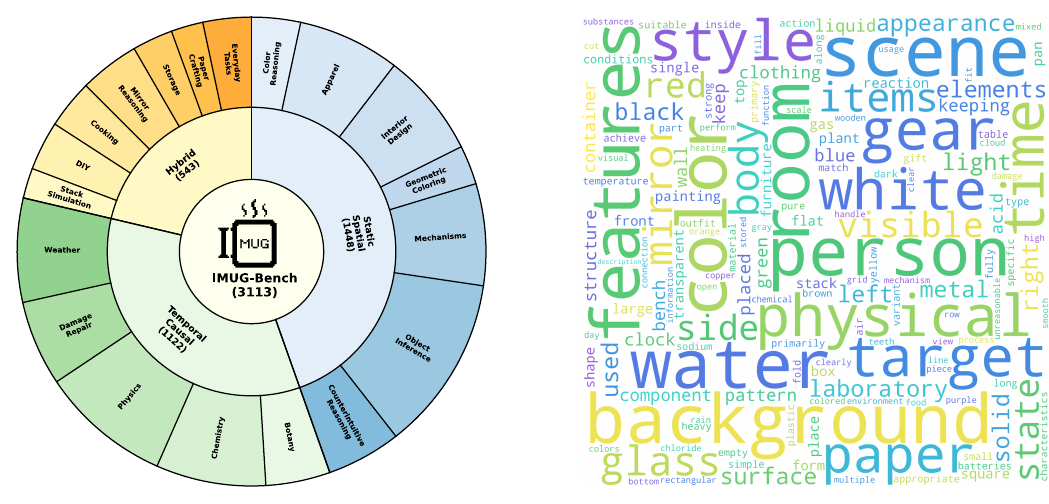}
    \caption{IMUG-Bench statistics and distribution. Left: pie-chart visualizations of the class/domain distribution of IMUG-Bench; Right: a word cloud summarizing the most frequent concepts.}
    \label{fig:world_cloud}
\end{figure}

Figure~\ref{fig:world_cloud} provides a high-level view of the benchmark distribution. The left side shows pie-chart visualizations of the class/domain composition of IMUG-Bench, and the right side shows a word cloud of the most frequent concepts, which together help illustrate the overall coverage and topic diversity of the benchmark.

Table~\ref{tab:dataset_stats} provides a detailed breakdown of the sample counts across each domain and task within IMUG-Bench, alongside their respective modality distributions (modality sequence) and interaction turn statistics.

\begin{longtable}{@{} p{0.14\linewidth} >{\centering\arraybackslash}p{0.5\linewidth} c @{\hspace{2.5em}} p{0.24\linewidth} @{}}
  \caption{Data composition and distribution summary table of IMUG-Bench. The numbers following Domain and Task indicate the number of samples within that partition. Turns represent the corresponding number of interaction turns under this interleaved sequence. Modality Sequence illustrates the possible alternating sequence of text--image output modalities under each task, where T represents text and I represents image.}
  \label{tab:dataset_stats} \\
  \toprule
  \textbf{Domain} & \textbf{Task} & \textbf{Turns} & \textbf{Modality} \newline \textbf{Sequence} \\
  \midrule
  \endfirsthead
  \multicolumn{4}{c}{{\tablename\ \thetable{} -- continued from previous page}} \\
  \toprule
  \textbf{Domain} & \textbf{Task} & \textbf{Turns} & \textbf{Modality} \newline \textbf{Sequence} \\
  \midrule
  \endhead
  \midrule
  \multicolumn{4}{r}{{Continued on next page}} \\
  \endfoot
  \bottomrule
  \endlastfoot
  \multirow[t]{17}{=}{Apparel \newline (233)} & Background Replacement (21) & 4 & [T, \allowbreak I, \allowbreak T, \allowbreak I] \\
   &  & 4 & [I, \allowbreak T, \allowbreak T, \allowbreak I] \\
   & Facial Feature Modification (18) & 2 & [I, \allowbreak T] \\
   &  & 2 & [T, \allowbreak I] \\
   & Full Body Extension (26) & 5 & [I, \allowbreak T, \allowbreak I, \allowbreak T, \allowbreak I] \\
   &  & 5 & [T, \allowbreak I, \allowbreak I, \allowbreak T, \allowbreak I] \\
   & Garment Replacement (26) & 3 & [I, \allowbreak T, \allowbreak I] \\
   &  & 3 & [T, \allowbreak I, \allowbreak I] \\
   & Gender Swapping (47) & 4 & [T, \allowbreak I, \allowbreak I, \allowbreak I] \\
   &  & 4 & [I, \allowbreak T, \allowbreak I, \allowbreak I] \\
   & Multi Person Portrait (27) & 2 & [I, \allowbreak T] \\
   &  & 3 & [T, \allowbreak T, \allowbreak I] \\
   &  & 2 & [T, \allowbreak I] \\
   & Pose Alteration (48) & 3 & [T, \allowbreak I, \allowbreak I] \\
   &  & 3 & [I, \allowbreak T, \allowbreak I] \\
   & Scene Adaptation (20) & 4 & [T, \allowbreak I, \allowbreak T, \allowbreak I] \\
   &  & 4 & [I, \allowbreak T, \allowbreak T, \allowbreak I] \\
  \midrule
  \multirow[t]{8}{=}{Botany \newline (132)} & Abiotic Stress (20) & 3 & [I, \allowbreak T, \allowbreak I] \\
   & Microbial Culture (15) & 3 & [I, \allowbreak I, \allowbreak T] \\
   & Natural Growth (25) & 3 & [I, \allowbreak T, \allowbreak I] \\
   &  & 3 & [I, \allowbreak I, \allowbreak T] \\
   & Nutrient Deficiency (15) & 3 & [I, \allowbreak T, \allowbreak I] \\
   & Plant Identification (22) & 3 & [T, \allowbreak T, \allowbreak T] \\
   & Tropisms (35) & 3 & [I, \allowbreak T, \allowbreak I] \\
   &  & 3 & [I, \allowbreak I, \allowbreak T] \\
  \midrule
  \multirow[t]{36}{=}{Chemistry \newline (254)} & Acid Reaction (21) & 4 & [I, \allowbreak T, \allowbreak I, \allowbreak T] \\
   &  & 4 & [T, \allowbreak I, \allowbreak I, \allowbreak T] \\
   &  & 4 & [I, \allowbreak T, \allowbreak I, \allowbreak I] \\
   &  & 4 & [T, \allowbreak I, \allowbreak I, \allowbreak I] \\
   & Apparatus Usage (29) & 4 & [I, \allowbreak T, \allowbreak I, \allowbreak T] \\
   &  & 4 & [T, \allowbreak I, \allowbreak I, \allowbreak T] \\
   & Ball Stick Model (12) & 4 & [I, \allowbreak T, \allowbreak I, \allowbreak I] \\
   &  & 4 & [T, \allowbreak I, \allowbreak I, \allowbreak I] \\
   & Combustion Flame (28) & 3 & [T, \allowbreak I, \allowbreak I] \\
   &  & 3 & [I, \allowbreak T, \allowbreak I] \\
   &  & 5 & [I, \allowbreak T, \allowbreak I, \allowbreak I, \allowbreak T] \\
   &  & 5 & [T, \allowbreak I, \allowbreak I, \allowbreak I, \allowbreak T] \\
   & Crystallization (7) & 4 & [T, \allowbreak I, \allowbreak T, \allowbreak I] \\
   &  & 4 & [I, \allowbreak T, \allowbreak T, \allowbreak I] \\
   & Daily Chemistry (22) & 5 & [I, \allowbreak T, \allowbreak T, \allowbreak I, \allowbreak T] \\
   &  & 5 & [T, \allowbreak I, \allowbreak T, \allowbreak I, \allowbreak T] \\
   & Gas Preparation (11) & 4 & [I, \allowbreak T, \allowbreak T, \allowbreak T] \\
   &  & 4 & [T, \allowbreak I, \allowbreak T, \allowbreak T] \\
   & Lab Accident (42) & 5 & [I, \allowbreak T, \allowbreak I, \allowbreak T, \allowbreak T] \\
   &  & 5 & [T, \allowbreak I, \allowbreak I, \allowbreak T, \allowbreak I] \\
   &  & 5 & [T, \allowbreak I, \allowbreak I, \allowbreak T, \allowbreak T] \\
   &  & 5 & [I, \allowbreak T, \allowbreak I, \allowbreak T, \allowbreak I] \\
   & Metal Displacement (14) & 4 & [I, \allowbreak T, \allowbreak T, \allowbreak I] \\
   &  & 4 & [T, \allowbreak I, \allowbreak T, \allowbreak I] \\
   &  & 4 & [I, \allowbreak T, \allowbreak I, \allowbreak T] \\
   &  & 4 & [T, \allowbreak I, \allowbreak I, \allowbreak T] \\
   & Ph Measurement (10) & 4 & [I, \allowbreak T, \allowbreak T, \allowbreak I] \\
   &  & 4 & [T, \allowbreak I, \allowbreak T, \allowbreak I] \\
   & Precipitation (20) & 4 & [I, \allowbreak T, \allowbreak I, \allowbreak T] \\
   &  & 4 & [I, \allowbreak T, \allowbreak T, \allowbreak I] \\
   &  & 4 & [T, \allowbreak I, \allowbreak T, \allowbreak I] \\
   &  & 4 & [T, \allowbreak I, \allowbreak I, \allowbreak T] \\
   & Specific Color Reaction (38) & 3 & [T, \allowbreak I, \allowbreak I] \\
   &  & 3 & [I, \allowbreak T, \allowbreak T] \\
   &  & 3 & [T, \allowbreak I, \allowbreak T] \\
   &  & 3 & [I, \allowbreak T, \allowbreak I] \\
  \midrule
  \multirow[t]{3}{=}{Color \newline Reasoning \newline (90)} & Additive Lighting Mixing (30) & 5 & [I, \allowbreak I, \allowbreak T, \allowbreak I, \allowbreak T] \\
   & Liquid Color Mixing (30) & 5 & [T, \allowbreak I, \allowbreak I, \allowbreak T, \allowbreak T] \\
   & Subtractive Color Mixing (30) & 4 & [I, \allowbreak I, \allowbreak T, \allowbreak T] \\
  \midrule
  \multirow[t]{3}{=}{Cooking \newline (90)} & Cooking Workflow (30) & 3 & [T, \allowbreak T, \allowbreak I] \\
   & Food Frying Compare (30) & 3 & [I, \allowbreak I, \allowbreak T] \\
   & Food Frying Flip (30) & 6 & [I, \allowbreak I, \allowbreak T, \allowbreak I, \allowbreak I, \allowbreak T] \\
  \midrule
  \multirow[t]{4}{=}{Counterin-\newline tuitive \newline Reasoning \newline (159)} & Action Relation Mismatch (40) & 3 & [I, \allowbreak T, \allowbreak I] \\
   & Compound Mismatch (39) & 4 & [T, \allowbreak I, \allowbreak T, \allowbreak I] \\
   & Object Attribute Mismatch (40) & 3 & [I, \allowbreak T, \allowbreak I] \\
   & Scene Object Mismatch (40) & 3 & [I, \allowbreak T, \allowbreak I] \\
  \midrule
  \multirow[t]{5}{=}{DIY \newline (91)} & Gift Wrapping Logic (31) & 6 & [T, \allowbreak I, \allowbreak T, \allowbreak I, \allowbreak T, \allowbreak I] \\
   & Rubber Stamp Carving (30) & 4 & [I, \allowbreak T, \allowbreak I, \allowbreak I] \\
   & Wax Seal Process (30) & 4 & [I, \allowbreak I, \allowbreak I, \allowbreak I] \\
   &  & 4 & [I, \allowbreak I, \allowbreak I, \allowbreak T] \\
   &  & 4 & [I, \allowbreak I, \allowbreak T, \allowbreak I] \\
  \midrule
  \multirow[t]{5}{=}{Damage \newline Repair \newline (188)} & Disaster Analysis (38) & 6 & [T, \allowbreak T, \allowbreak I, \allowbreak T, \allowbreak T, \allowbreak I] \\
   &  & 3 & [T, \allowbreak T, \allowbreak I] \\
   & Item Repair (50) & 4 & [T, \allowbreak T, \allowbreak T, \allowbreak I] \\
   & Relic Restoration (50) & 3 & [I, \allowbreak T, \allowbreak I] \\
   & Rust Restoration (50) & 6 & [I, \allowbreak T, \allowbreak I, \allowbreak T, \allowbreak I, \allowbreak T] \\
  \midrule
  \multirow[t]{3}{=}{Everyday \newline Tasks \newline (89)} & Battery Installation (30) & 4 & [I, \allowbreak T, \allowbreak I, \allowbreak T] \\
   & Glass Cleaning (29) & 4 & [I, \allowbreak T, \allowbreak I, \allowbreak T] \\
   & Wound Treatment (30) & 3 & [T, \allowbreak I, \allowbreak I] \\
  \midrule
  \multirow[t]{6}{=}{Geometric \newline Coloring \newline (64)} & Cubes Coloring (10) & 3 & [T, \allowbreak I, \allowbreak I] \\
   & Cubes Completion (7) & 3 & [T, \allowbreak I, \allowbreak T] \\
   & Cubes Synthesis (7) & 6 & [T, \allowbreak I, \allowbreak T, \allowbreak T, \allowbreak I, \allowbreak I] \\
   & Grid Coloring (40) & 4 & [I, \allowbreak I, \allowbreak T, \allowbreak I] \\
   &  & 5 & [I, \allowbreak I, \allowbreak T, \allowbreak I, \allowbreak T] \\
   &  & 5 & [I, \allowbreak T, \allowbreak I, \allowbreak T, \allowbreak I] \\
  \midrule
  \multirow[t]{6}{=}{Interior \newline Design \newline (240)} & Artwork Addition (40) & 6 & [I, \allowbreak T, \allowbreak I, \allowbreak I, \allowbreak I, \allowbreak T] \\
   & Furniture Arrangement (40) & 5 & [I, \allowbreak T, \allowbreak I, \allowbreak I, \allowbreak T] \\
   & Global Style Modification (40) & 3 & [I, \allowbreak T, \allowbreak I] \\
   & Local Style Modification (40) & 3 & [T, \allowbreak T, \allowbreak I] \\
   & Room Renovation (40) & 2 & [T, \allowbreak I] \\
   & Scene Completion (40) & 2 & [T, \allowbreak I] \\
  \midrule
  \multirow[t]{9}{=}{Mechan-\newline isms \newline (232)} & Gear Completion (40) & 2 & [T, \allowbreak I] \\
   & Gear Meshing (40) & 3 & [T, \allowbreak T, \allowbreak I] \\
   & Object Pairing (30) & 2 & [T, \allowbreak I] \\
   & Rotational Mechanism Edit (31) & 2 & [I, \allowbreak I] \\
   & Rotational Mechanism Generation (31) & 3 & [T, \allowbreak I, \allowbreak I] \\
   & Sliding Mechanism Edit (30) & 2 & [I, \allowbreak I] \\
   & Sliding Mechanism Generation (30) & 3 & [T, \allowbreak I, \allowbreak T] \\
   &  & 2 & [T, \allowbreak I] \\
   &  & 3 & [T, \allowbreak I, \allowbreak I] \\
  \pagebreak
  \multirow[t]{4}{=}{Mirror \newline Reasoning \newline (120)} & Mirror Action Editing (40) & 3 & [I, \allowbreak T, \allowbreak I] \\
   & Mirror Clock Person Composite (40) & 4 & [I, \allowbreak I, \allowbreak T, \allowbreak T] \\
   & Mirror Clock Reading (40) & 2 & [T, \allowbreak I] \\
   &  & 3 & [T, \allowbreak I, \allowbreak T] \\
  \multirow[t]{16}{=}{Object \newline Inference \newline (430)} & Consumer Electronics (55) & 5 & [I, \allowbreak I, \allowbreak I, \allowbreak T, \allowbreak T] \\
   &  & 6 & [I, \allowbreak I, \allowbreak I, \allowbreak T, \allowbreak I, \allowbreak T] \\
   & Daily Essentials (11) & 5 & [I, \allowbreak I, \allowbreak I, \allowbreak T, \allowbreak T] \\
   &  & 6 & [I, \allowbreak I, \allowbreak I, \allowbreak T, \allowbreak I, \allowbreak T] \\
   & Furniture And Interior (77) & 5 & [I, \allowbreak I, \allowbreak I, \allowbreak T, \allowbreak T] \\
   &  & 6 & [I, \allowbreak I, \allowbreak I, \allowbreak T, \allowbreak I, \allowbreak T] \\
   & Hardware Tools (55) & 5 & [I, \allowbreak I, \allowbreak I, \allowbreak T, \allowbreak T] \\
   &  & 6 & [I, \allowbreak I, \allowbreak I, \allowbreak T, \allowbreak I, \allowbreak T] \\
   & Kitchenware And Appliances (67) & 5 & [I, \allowbreak I, \allowbreak I, \allowbreak T, \allowbreak T] \\
   &  & 6 & [I, \allowbreak I, \allowbreak I, \allowbreak T, \allowbreak I, \allowbreak T] \\
   & Recreation And Instruments (55) & 5 & [I, \allowbreak I, \allowbreak I, \allowbreak T, \allowbreak T] \\
   &  & 6 & [I, \allowbreak I, \allowbreak I, \allowbreak T, \allowbreak I, \allowbreak T] \\
   & Scientific And Professional Equip (55) & 5 & [I, \allowbreak I, \allowbreak I, \allowbreak T, \allowbreak T] \\
   &  & 6 & [I, \allowbreak I, \allowbreak I, \allowbreak T, \allowbreak I, \allowbreak T] \\
   & Stationery And Office (55) & 5 & [I, \allowbreak I, \allowbreak I, \allowbreak T, \allowbreak T] \\
   &  & 6 & [I, \allowbreak I, \allowbreak I, \allowbreak T, \allowbreak I, \allowbreak T] \\
  \midrule
  \multirow[t]{3}{=}{Paper \newline Crafting \newline (45)} & Cutting (15) & 5 & [I, \allowbreak T, \allowbreak I, \allowbreak T, \allowbreak I] \\
   & Origami Basic Geometry (15) & 4 & [I, \allowbreak T, \allowbreak I, \allowbreak I] \\
   & Origami Sequential Logic (15) & 6 & [I, \allowbreak T, \allowbreak I, \allowbreak I, \allowbreak T, \allowbreak T] \\
  \midrule
  \multirow[t]{12}{=}{Physics \newline (310)} & Clock Operation (40) & 4 & [I, \allowbreak I, \allowbreak T, \allowbreak T] \\
   & Deformation (30) & 3 & [T, \allowbreak I, \allowbreak I] \\
   & Diffusion Tyndall (30) & 4 & [I, \allowbreak I, \allowbreak T, \allowbreak I] \\
   & Magnetism Advanced (30) & 3 & [I, \allowbreak T, \allowbreak I] \\
   & Magnetism Basic (30) & 3 & [I, \allowbreak T, \allowbreak I] \\
   & Magnetism Compare (30) & 4 & [I, \allowbreak I, \allowbreak T, \allowbreak T] \\
   & Mass Measurement (30) & 4 & [T, \allowbreak I, \allowbreak I, \allowbreak I] \\
   & Refraction Multi Layer (30) & 4 & [I, \allowbreak T, \allowbreak T, \allowbreak I] \\
   & Refraction Single (30) & 4 & [T, \allowbreak I, \allowbreak T, \allowbreak I] \\
   &  & 2 & [I, \allowbreak I] \\
   &  & 3 & [T, \allowbreak I, \allowbreak I] \\
   & State Change (30) & 6 & [T, \allowbreak I, \allowbreak T, \allowbreak I, \allowbreak I, \allowbreak T] \\
  \midrule
  \multirow[t]{2}{=}{Stack \newline Simulation \newline (40)} & Stack Math Expression (20) & 5 & [I, \allowbreak T, \allowbreak I, \allowbreak T, \allowbreak T] \\
   & Stack Word Formation (20) & 5 & [I, \allowbreak T, \allowbreak I, \allowbreak T, \allowbreak I] \newline \\
  \midrule
  Storage \newline (68) & Elastic Storage (68) & 4 & [I, \allowbreak I, \allowbreak T, \allowbreak T] \\
  \midrule
  \multirow[t]{10}{=}{Weather \newline (238)} & Cloud Dynamics (30) & 2 & [I, \allowbreak I] \\
   & Cloud State Evolution (27) & 3 & [T, \allowbreak T, \allowbreak I] \\
   &  & 2 & [T, \allowbreak I] \\
   & Cumulative Effects (61) & 3 & [I, \allowbreak T, \allowbreak I] \\
   &  & 2 & [I, \allowbreak I] \\
   & Rain Gradient Visuals (30) & 2 & [I, \allowbreak I] \\
   &  & 2 & [T, \allowbreak I] \\
   & Solar Lighting (30) & 5 & [I, \allowbreak I, \allowbreak I, \allowbreak I, \allowbreak I] \\
   & Visibility Obscuration (30) & 4 & [I, \allowbreak I, \allowbreak T, \allowbreak T] \\
   & Wind Force Modification (30) & 2 & [T, \allowbreak I] \\
\end{longtable}

\section{Additional Experimental Results}
\label{sec:appendix_additional_results}

\subsection{Radar Charts of Model Performance}
\label{sec:appendix_radar_results}

\begin{figure}[H]
    \centering
    \includegraphics[width=0.9\linewidth]{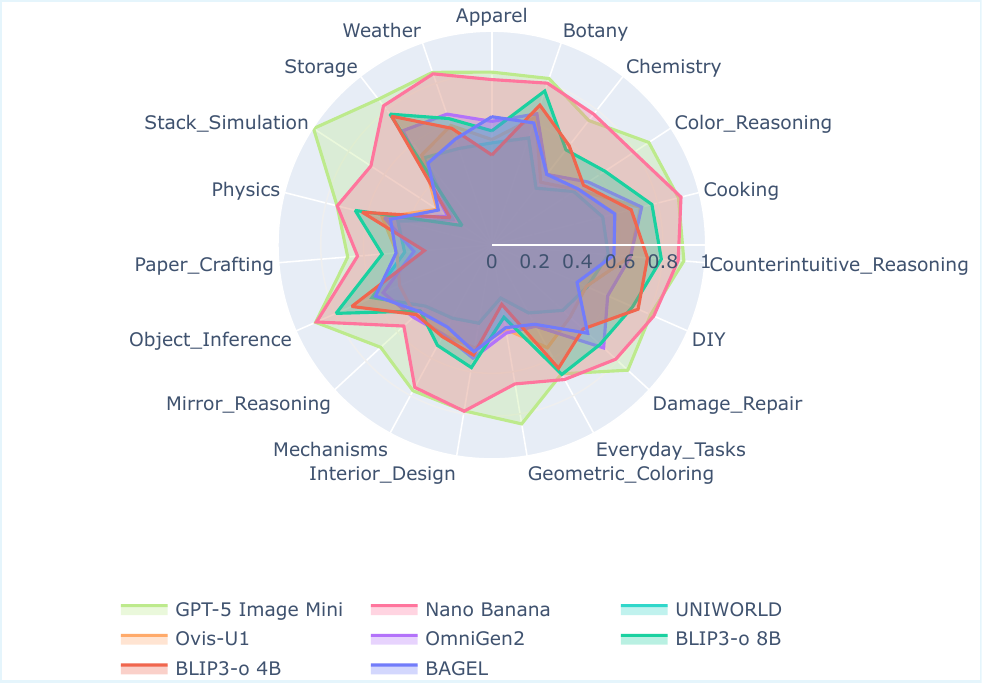}
    \caption{Scores of all modalities for each model across all tasks.}
    \label{fig:c1}
\end{figure}

\begin{figure}[H]
    \centering
    \includegraphics[width=0.9\linewidth]{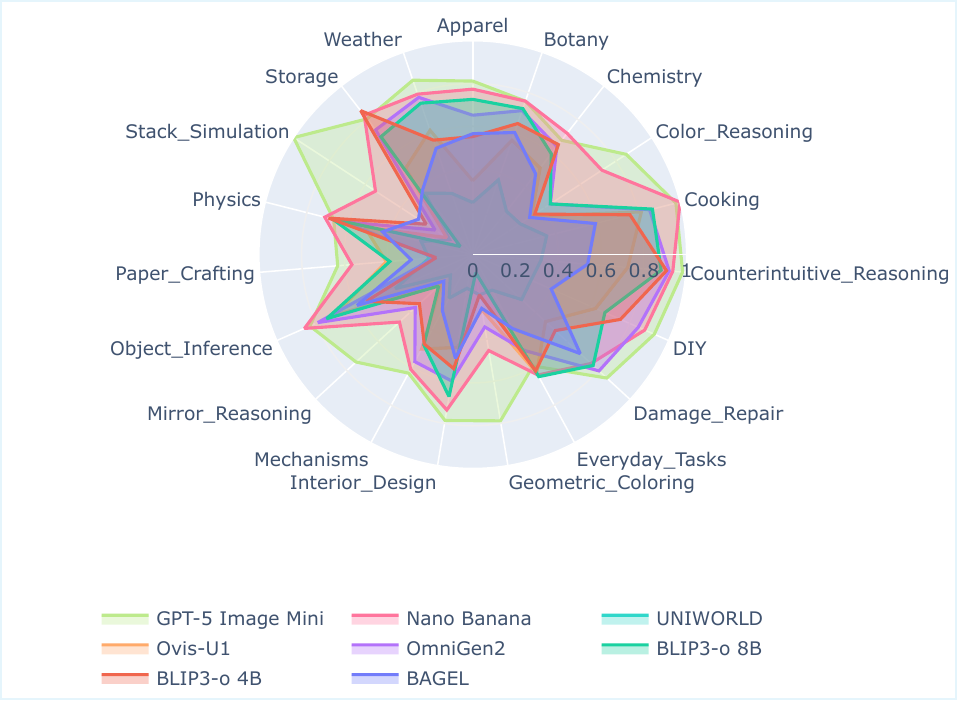}
    \caption{Scores of text-only modalities for each model across all tasks.}
    \label{fig:c2}
\end{figure}

\begin{figure}[H]
    \centering
    \includegraphics[width=0.9\linewidth]{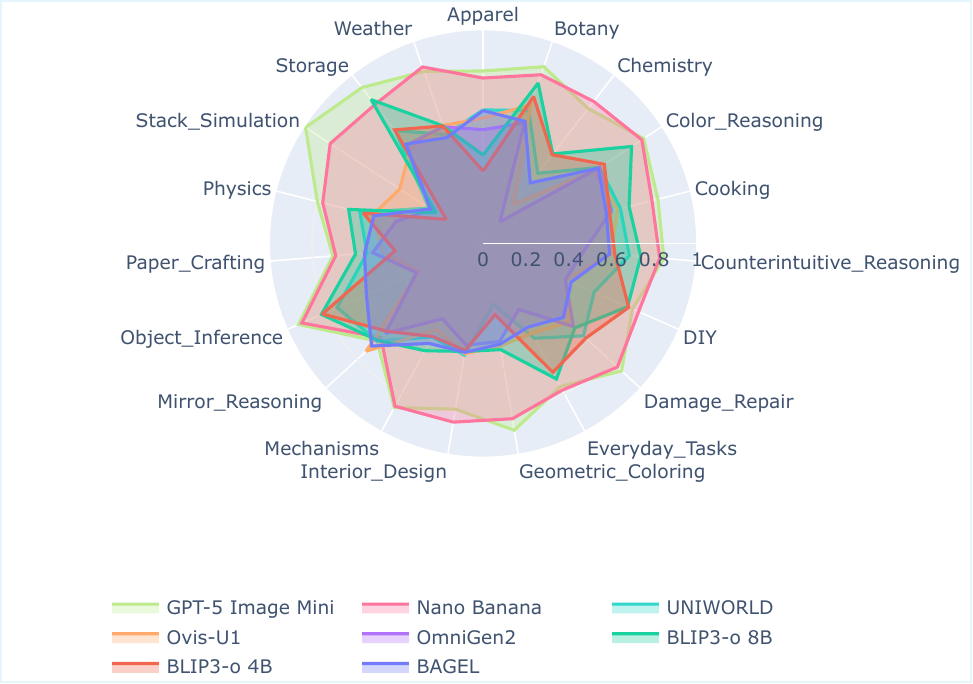}
    \caption{Scores of image-only modalities for each model across all tasks.}
    \label{fig:c3}
\end{figure}

This subsection visualizes the task-specific scores for all tested models via radar charts. Specifically, Figs.~\ref{fig:c1}, \ref{fig:c2}, and \ref{fig:c3} display the results for all modalities, text-only modalities, and image-only modalities.

In these radar charts, lines of different colors represent the performance scores of various models. Based on the specific trends of the curves, several significant common characteristics across models can be identified. For the text modality, the six open-source UMMs generally perform poorly in Mirror Reasoning and Color Reasoning tasks, while showing relatively superior performance in tasks such as Cooking and Counterintuitive Reasoning. Notably, some open-source models achieve scores closely approaching those of closed-source UMMs in a wide range of tasks. Regarding the image modality, all six open-source UMMs exhibit a relative deficiency in the Chemistry task, a weakness not observed in the two closed-source models. Furthermore, the scores of open-source UMMs across all modalities remain consistently low in Stack Simulation, Paper Crafting, and Geometric Coloring tasks.

These results reveal several key insights. First, the performance gap between open-source UMMs and closed-source models is narrower in understanding tasks compared to generation tasks; specifically, in relatively daily or simple scenarios, certain open-source UMMs have approached the performance level of closed-source models, though a significant gap remains in complex logical reasoning. Second, the generative capabilities of open-source UMMs in highly specialized domains---such as chemical substances, laboratory equipment, and stack simulation---require further enhancement, as they often struggle to accurately render professional domain-specific diagrams requested by users. Finally, for tasks demanding the highest level of integrated spatio-temporal reasoning and generation across all modalities, all open-source UMMs show substantial room for improvement. Even in extremely complex tasks like Paper Crafting (paper folding and cutting) and Mirror Reasoning (rendering clocks and human actions in mirrors), closed-source models also exhibit noticeably lower scores compared to their performance in other tasks.

\vspace{20pt}
\newpage
\vspace{20pt}
\FloatBarrier
\newpage

\subsection{Score Trends over Interaction Turns}
\label{sec:appendix_turn_trends}

\begin{figure}[t]
    \centering
    \includegraphics[width=1\linewidth]{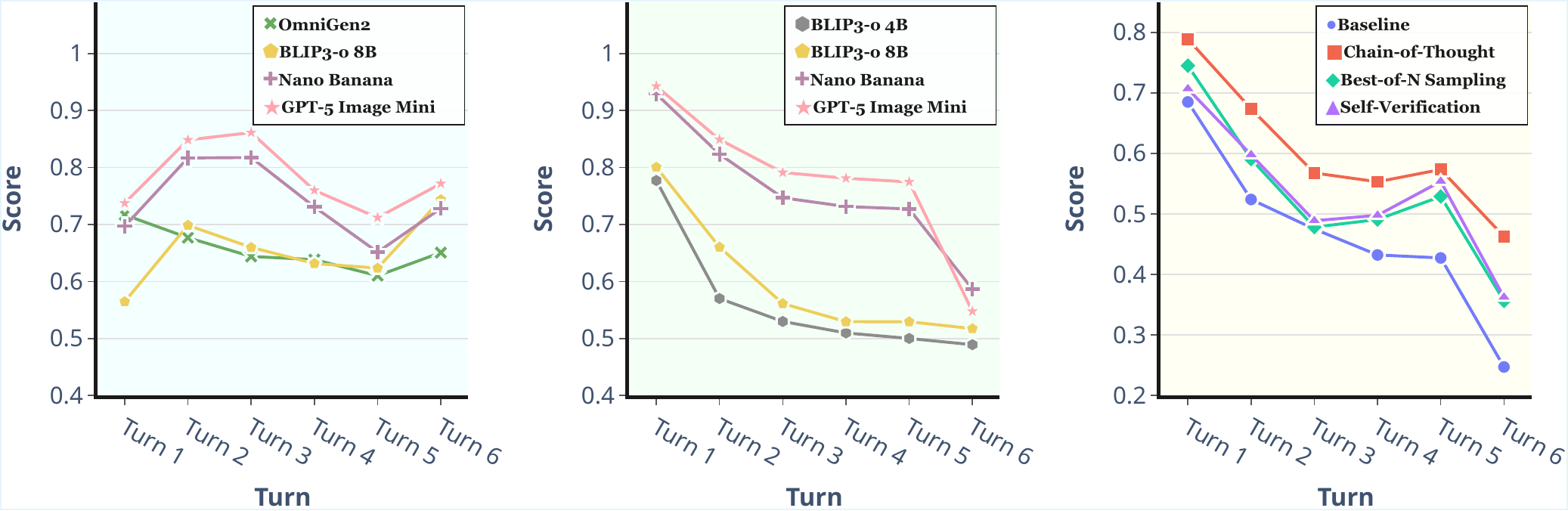}
    \caption{Score trends over interaction turns. Left: text-modality question scores of the top-4 models. Middle: image-modality question scores of the top-4 models. Right: BAGEL baseline compared with three test-time scaling strategies across turns.}
    \label{fig:exposurebias}
\end{figure}

Figure~\ref{fig:exposurebias} provides a more intuitive visualization of these trends, including the comparison between the BAGEL baseline and the three test-time scaling strategies.

\subsection{Supplementary Results on Static-MCQ and Dynamic-MCQ}
\label{sec:appendix_sup_of_mcq}

Table~\ref{tab:static_dynamic_mcq} presents the detailed scores of all evaluated models on Static-MCQ and Dynamic-MCQ. These results provide numerical evidence and further support the corresponding findings discussed in Section~\ref{main_result}.

\begin{table}[tb]
  \caption{Comparison of model performance on Static-MCQ and Dynamic-MCQ.}
  \label{tab:static_dynamic_mcq}
  \centering
  
  \newcolumntype{M}[1]{>{\centering\arraybackslash}m{#1}}
  \newcolumntype{L}[1]{>{\raggedright\arraybackslash}m{#1}} 

  \begin{threeparttable}
  \begin{tabular}{@{}M{0.23\linewidth} M{0.18\linewidth} M{0.18\linewidth}@{}}
    \toprule
    \textbf{Model Name} & \textbf{Static-MCQ} & \textbf{Dynamic-MCQ} \\
    \midrule

    GPT-5 Image Mini\tnote{1} & 83.5 & 50.6 \\
    Nano Banana\tnote{2} & 80.2 & 43.4 \\
    \midrule
    BAGEL\tnote{3} & 52.7 & 25.3 \\
    BLIP3-o 4B\tnote{4} & 58.8 & 46.4 \\
    BLIP3-o 8B\tnote{4} & 67.3 & 47.0 \\
    OmniGen2\tnote{5} & 72.3 & 30.0 \\
    Ovis-U1\tnote{6} & 56.1 & 27.8 \\
    UNIWORLD-V1\tnote{7} & 29.2 & 20.9 \\

    \bottomrule
  \end{tabular}

\begin{tablenotes}[flushleft]
  \footnotesize
  \item[] [1]~\cite{singh2025openai}; 
         [2]~\cite{comanici2025gemini}; 
         [3]~\cite{deng2025emerging}; 
         [4]~\cite{chen2025blip3}; 
         [5]~\cite{wu2025omnigen2}; 
         [6]~\cite{wang2025ovis}; 
         [7]~\cite{lin2025uniworld}.
\end{tablenotes}
  \end{threeparttable}
\end{table}

\section{Human Validation}
\label{app:human_validation}
To validate the reliability of the VLM-as-a-judge evaluation, we use the Pearson correlation coefficient to measure the agreement between our automated scores and human ratings.
We collect outputs from two closed-source models, and from open-source models we select three models with the highest, medium, and lowest scores.
To ensure representativeness, for each model we sample one task-sample response from each task randomly.
Based on this setup, we further conduct the human validation experiment on five representative models.
We sample the inference results of five models---GPT-5 Image Mini, Nano Banana, BLIP3-o 8B, OmniGen2, and UNIWORLD-V1---for manual re-scoring, and compute the Pearson correlation coefficient between human ratings and VLM-as-a-judge scores; the results are shown in Table~\ref{tab:model_scores}.
The Pearson correlations exceed 0.72 for all row-wise (model-level) and column-wise (class-level) averages, with an overall mean above 0.8, indicating that our evaluation method closely matches human scoring and supporting the effectiveness of VLM-as-a-judge. This strong alignment further provides a necessary foundation for scalable and automated evaluation in large-scale settings.

\begin{table}[tb]
  \caption{Human validation of VLM-as-a-judge scoring. We report Pearson correlations between automated and two-person human scores for Img. and Dynamic-MCQ across classes and on average.}
  \label{tab:model_scores}
  \centering

  \newcolumntype{M}[1]{>{\centering\arraybackslash}m{#1}}
  \newcolumntype{Y}{>{\centering\arraybackslash}X}
  \newcolumntype{L}[1]{>{\raggedright\arraybackslash}m{#1}} 

  \begin{threeparttable}
  \begin{tabularx}{\linewidth}{@{}L{0.2\linewidth} *{7}{Y}@{}}
    \toprule
    \multirow{2}{*}{\textbf{Model Name}} &
    \multicolumn{2}{c}{\textbf{St.\&Sp.}} &
    \multicolumn{2}{c}{\textbf{Te.\&Ca.}} &
    \multicolumn{2}{c}{\textbf{Hybrid}} &
    \multirow{2}{*}{\textbf{Avg}} \\
    \cmidrule(lr){2-3}\cmidrule(lr){4-5}\cmidrule(lr){6-7}
    & \textbf{Img.} & \textbf{Dynamic} & \textbf{Img.} & \textbf{Dynamic} & \textbf{Img.} & \textbf{Dynamic} & \\
    \midrule
    GPT-5 Image Mini\tnote{1} & 0.725 & 0.813 & 0.662 & 0.839 & 0.771 & 0.884 & 0.782 \\
    Nano Banana\tnote{2} & 0.839 & 0.926 & 0.781 & 0.816 & 0.706 & 0.855 & 0.821 \\
    \midrule
    BLIP3-o 8B\tnote{3} & 0.759 & 0.821 & 0.702 & 0.916 & 0.735 & 0.830 & 0.794 \\
    OmniGen2\tnote{4} & 0.693 & 0.802 & 0.775 & 0.860 & 0.740 & 0.831 & 0.784 \\
    UNIWORLD-V1\tnote{5} & 0.837 & 0.861 & 0.878 & 0.871 & 0.693 & 0.907 & 0.841 \\
    \midrule
    \textbf{Average} & 0.771 & 0.845 & 0.760 & 0.860 & 0.729 & 0.861 & 0.804 \\
    \bottomrule
  \end{tabularx}

  \begin{tablenotes}[flushleft]
    \footnotesize
    \item[] [1]~\cite{singh2025openai}; [2]~\cite{comanici2025gemini}; [3]~\cite{chen2025blip3}; [4]~\cite{wu2025omnigen2}; [5]~\cite{lin2025uniworld}.
  \end{tablenotes}
  \end{threeparttable}
\end{table}

\section{Prompts}

In this section, we list the prompts employed throughout the construction, testing, and scoring pipelines of IMUG-Bench.

\subsection{Prompts for Benchmark Construction}
\label{sec:appendix_prompts_question_filling}

The following prompt is employed for task sample filling and construction. During the construction process, this prompt is immediately followed by the question template and keyword sequences to be filled; if images are involved, they are provided as input simultaneously.

\begin{tcolorbox}[
  breakable,
  colback=white,
  colframe=black,
  boxrule=0.8pt,
  arc=0pt,
  left=2mm,right=2mm,top=1mm,bottom=1mm
]
\footnotesize
\begin{Verbatim}[
  breaklines=true,
  breakanywhere=true,
  showspaces=false,
  showtabs=false,
  breaksymbolleft={},
  breaksymbolright={},
  breaksymbol={ }
]
[System Prompt]
1. Role
You are an expert Dataset Generator for a Multimodal Benchmark.

2. Task
Your task is to instantiate test samples by combining Paradigm Templates with Blueprint Variables.

3. Execution Logic
3.1 Analyze Context
Specify every keyword from the Blueprint.

3.2 Fill Placeholders
Replace all <...> in the "prompt_template" with Context Variables.
Constraint: Do NOT rewrite the template text outside brackets.
CRITICAL: You must reconstruct the sentence to ensure it sounds like it was written by a native English speaker.

3.3 Construct Options
Generate 4 specific options (A-D).
Option E must be "None of the above".
Randomization: Randomize the position of the correct option.

3.4 Construct Evaluation Points
Fill and expand specific checks into detailed points. Please confirm that all evaluation-points are clear and unambiguous.

4. Output Format
Output ONLY valid JSON. No markdown.

{
    "sample_id": <Int>,
    "category": "<String, Subdomain>",
    "total_tasks": <Int>,
    "tasks": [
        {
            "turn": <Int>,
            "modality": "text",
            "input": [
                {
                    "text": "<String: English Question + \nOptions:\nA. ...\nB. ...\nC. ...\nD. ...\nE. None of the above.>"
                }
            ],
            "output": {
                "answer": "<String: Correct Letter or '<DYNAMIC>'>"
            }
        }
    ]
}

5. Critical Constraints
5.1 Logic Consistency
Ensure strict physical and logical consistency.

5.2 Anti-Leakage
Current question stem must NOT reveal the current answer.
Subsequent question stems must NOT reveal answers to previous turns.

5.3 Conciseness
Keep descriptions brief and accurate.

[Question Template]
...

[Keyword-Value Pairs]
...

[Image Sequence (if applicable)]
...
\end{Verbatim}
\end{tcolorbox}

\subsection{Prompts for Model Evaluation}
\label{sec:appendix_prompts_evaluation}

Below is the system prompt provided to the model during the IMUG-Bench evaluation process. This prompt is immediately followed by the main body of the test questions.

\begin{tcolorbox}[
  breakable,
  colback=white,
  colframe=black,
  boxrule=0.8pt,
  arc=0pt,
  left=2mm,right=2mm,top=1mm,bottom=1mm
]
\footnotesize
\begin{Verbatim}[
  breaklines=true,
  breakanywhere=true,
  showspaces=false,
  showtabs=false,
  breaksymbolleft={},
  breaksymbolright={},
  breaksymbol={ }
]
[System Prompt]
1. Role
You are an AI assistant participating in a multimodal benchmark evaluation.

2. Task
This is a multi-turn multimodal task. You must synthesize all history information (including all prompt texts, input images, and your own previous text/image outputs) to determine your response strategy. Inputs may contain text and single or multiple images.

3. Rules
3.1 Output Modality Constraints
Your response for each turn must be strictly EITHER TEXT OR IMAGE.
If text output is required, treat it as a Multiple Choice Question (MCQ).
If image output is required, treat it as a Generating Task.

3.2 MCQ Guidelines
All MCQs are indeterminate choice (can be single or multiple selection). Determine the number of correct options based on the prompt.
Output Format: Output ONLY a contiguous string of uppercase letters representing the correct options, with no separators or punctuation (e.g., 'A' or 'BCD'). Do NOT output option content or explanations.

3.3 Generating Task Guidelines
Generate one image based on all context.
Hint: Unless explicitly instructed to "generate a new image" or "modify a specific image", the images you generate should by default reference the most recent image in the context (whether input or previously generated by you) for further modification, or be re-created based on certain elements within it

[Multi-turn Tasks]
...
\end{Verbatim}
\end{tcolorbox}

\subsection{Prompts for Dynamic Answer Extraction}
\label{sec:appendix_prompts_dynamic_mcq}

Below is the prompt used to retrieve the standard answers for the dynamic MCQs. This prompt is input to the judge VLM along with the reference dialogue turns.

\begin{tcolorbox}[
  breakable,
  colback=white,
  colframe=black,
  boxrule=0.8pt,
  arc=0pt,
  left=2mm,right=2mm,top=1mm,bottom=1mm
]
\footnotesize
\begin{Verbatim}[
  breaklines=true,
  breakanywhere=true,
  showspaces=false,
  showtabs=false,
  breaksymbolleft={},
  breaksymbolright={},
  breaksymbol={ }
]
[System Prompt]
1. Role
You are a top-tier Multimodal Dialogue Benchmark Evaluation Expert. For this task, you will serve as the "Dynamic Answer Determination Assistant."

2. Task Background
The AI model being evaluated is participating in a multi-turn conversation where it generates or modifies images based on user instructions. The conversation has reached a "Dynamic Multiple-Choice Question (MCQ)" turn. This question is "dynamic" because its ground-truth answer is not fixed; instead, it depends entirely on the actual content generated by the model in previous reference turns (e.g., the number of objects drawn, the specific color of a background, etc.).

3. Task Objective
Your goal is to analyze the reference information and determine the correct standard answer for this context.

4. Provided Content
4.1 Reference Turns
This includes user instructions (Task Questions) from previous turns and the actual outputs (Model Response/Image) produced by the evaluated model. These serve as the only factual basis for determining the answer.

4.2 Final Dynamic Question
This is the last turn's MCQ. Note that the answer field may contain a <DYNAMIC> placeholder or partial answers (e.g., A+<DYNAMIC>, means 'A' must be a correct answer, and you need to identify the other correct items among the available options). You must identify all correct options (A, B, C, etc.) based on the actual generation results.

5. Execution Logic
5.1 Trace the Logic Chain
Carefully review the requirements in the Reference Turns and observe the final state of elements (color, position, quantity, style, etc.) in the generated images or text.

5.2 Map to Options
Compare the "factual state" from the reference outputs against each option in the Final Dynamic Question.

5.3 Eliminate Distractors
Even if an option is factually true in a general sense, if it does not match the specific content actually generated by the model, it must be considered incorrect.

6. Output Format
You must output strictly in JSON format. No additional text, explanations, or summaries are allowed. The structure must be as follows:

{
  "determined_answer": "A contiguous string of uppercase letters, e.g., 'A' or 'BCD'",
  "reasoning": "State your logic for determining the answer in detail. Mention which feature from which reference turn was used and how it maps to the chosen options."
}

[Target-Turn Dynamic-MCQ Question and Choices]
...

[Sequence of Reference Turn Q&As]
...
\end{Verbatim}
\end{tcolorbox}

\subsection{Prompts for Scoring}
\label{sec:appendix_prompts_image_assessment}

The following is the prompt used for image assessment. This prompt is provided to the judge VLM along with the sequence of reference images, the image to be evaluated, and their corresponding image sequence labels.

\begin{tcolorbox}[
  breakable,
  colback=white,
  colframe=black,
  boxrule=0.8pt,
  arc=0pt,
  left=2mm,right=2mm,top=1mm,bottom=1mm
]
\footnotesize
\begin{Verbatim}[
  breaklines=true,
  breakanywhere=true,
  showspaces=false,
  showtabs=false,
  breaksymbolleft={},
  breaksymbolright={},
  breaksymbol={ }
]
[System Prompt]
1. Role
You are a professional visual dialogue quality evaluation expert.

2. Task
Your task is to provide precise scoring for the image generation result of a specific turn within a "Multi-turn Multimodal Dialogue Benchmark." You need to judge whether the model-generated image meets the preset "evaluation-points" based on the instructions given in that turn, combined with historical reference information.

3. Task Background
This task is extracted from a continuous multi-turn dialogue between an AI and a user. During this process, the user continuously provides new instructions, requiring the AI to modify, evolve, maintain style, or supplement logic based on previous visual information (input images or images previously generated by the AI).

4. Provided Content
4.1 Reference Images (Image_1, Image_2, ..., Image_N-1)
These represent the key visual states that appeared earlier in the dialogue path. They may be original materials provided by the user (input images) or intermediate results generated by the AI (output images).

4.2 Target Image (Image_N, the last one)
The output generated by the model according to the latest instructions for the current turn.

5. Scoring Standards (0-5 Scale)
This is a fine-grained quality evaluation task, NOT a binary correctness check. Please score each evaluation point independently; scores must be integers. The score represents the "completeness" of the target image relative to a specific evaluation point. The more accurately the image completes the requirements and the higher the quality, the higher the score should be.

5.1 5 points (Flawless)
Must fully comply with the current task requirements.
Must maintain high "pixel-level" or "logic-level" consistency with the specified reference images regarding the required elements.
No visual distortions, physical logic errors, or feature loss.

5.2 4 points (Satisfactory)
The evaluation point is clearly satisfied, all required elements are present and recognizable.
Minor inconsistencies, simplifications, or small visual artifacts may exist, but they do not affect the overall fulfillment of this point.

5.3 3 points (Partial Achievement)
The evaluation point is generally satisfied at a semantic level.
Key elements are present, but there are noticeable visual inaccuracies, partial mismatches with reference images, or logical simplifications.
This score should be used when the intent is clear but execution is imperfect.

5.4 2 points (Significant Deviation)
Only partial features required by the evaluation point are retained, but the overall logic has shifted.
There are multiple significant visual errors or an obvious disconnect in feature inheritance from the reference images.

5.5 1 point (Trace Evidence)
There is a significant deviation from the evaluation point requirements, but the direction of completion shares a certain degree of similarity with the requirements, rather than being completely opposite or entirely unrelated.

5.6 0 points (Void)
The target image has absolutely no intersection with the content described in the evaluation (e.g., completely collapsed random noise or solid color blocks).

6. Output Format
Please output the evaluation results directly in JSON format, following the format below:
{
    "evaluation_results": [
        {
            "point_id": 1,
            "score": 5,
            "reason": "explaining the specific performance of the target image on this point."
        },
        {
            "point_id": 2,
            "score": 0,
            "reason": "explaining the specific cause for the total inconsistency on this point."
        }
    ]
}

[Target-Turn Question and Evaluation-Points]
...

[Sequence of Reference and Target Images]
...
\end{Verbatim}
\end{tcolorbox}

\end{document}

%% file: math_commands.tex
\usepackage{amsmath,amsfonts,bm}

\def\eqref#1{equation~\ref{#1}}

\def\1{\bm{1}}

\DeclareMathAlphabet{\mathsfit}{\encodingdefault}{\sfdefault}{m}{sl}
\SetMathAlphabet{\mathsfit}{bold}{\encodingdefault}{\sfdefault}{bx}{n}

